\documentclass[letterpaper]{article} 
\usepackage{aaai2026}  
\nocopyright
\usepackage[hyphens]{url}  
\usepackage{graphicx} 
\usepackage{natbib}  
\usepackage{caption} 
\usepackage{algorithm}
\usepackage{algorithmic}
\usepackage{amsmath}

\usepackage{microtype}
\usepackage{graphicx}
\usepackage{booktabs} 
\usepackage{multirow}
\usepackage[table,xcdraw]{xcolor}
\usepackage{arydshln}  
\usepackage{pifont}
\usepackage[table,xcdraw]{xcolor}
\usepackage{svg}

\usepackage{caption}
\usepackage{xcolor}

\usepackage{amsmath}
\usepackage{amssymb}
\usepackage{mathtools}
\usepackage{amsthm}
\usepackage{subcaption} 
\usepackage{arydshln}
\definecolor{light-gray0}{gray}{0.9}
\usepackage{newfloat}
\usepackage{listings}
\DeclareCaptionStyle{ruled}{labelfont=normalfont,labelsep=colon,strut=off} 
\floatstyle{ruled}
\newfloat{listing}{tb}{lst}{}
\floatname{listing}{Listing}

\usepackage{booktabs}

\title{TREND-10K: A Comprehensive Dataset for Next-Generation Video Quality Assessment Based on Preference-Driven Media}

\author {
    Ziheng~Jia\textsuperscript{\rm 1}, 
    Zicheng~Zhang\textsuperscript{\rm 2},
    Junqi~Zhang\textsuperscript{\rm 1},
    Jiaying~Qian\textsuperscript{\rm 1},
    Jiarui~Wang\textsuperscript{\rm 1},
    Yushuo~Zheng\textsuperscript{\rm 1},
    Xiongkuo~Min\textsuperscript{\rm 1}\thanks{Corresponding author.}
}
\affiliations {
    \textsuperscript{\rm 1}Shanghai Jiao Tong University\\
    \textsuperscript{\rm 2}Shanghai Artificial Intelligence Laboratory\\
    jzhws1@sjtu.edu.cn
}

\begin{document}

\maketitle

\begin{abstract}
The increasing prominence of short-video platforms, coupled with the advanced commercialization of AI-generated content~(AIGC) videos, has led to a shift in the types of video media \textbf{trend} consumed by users in their daily lives. Traditional user-generated content (UGC) is gradually being replaced by professional short dramas and AIGC entertainment. Consequently, VQA for contemporary media content has become increasingly important. This requires a unified evaluation framework that can handle diverse video content and evolving media trends. In this context, we introduce \textbf{TREND-10K}, a next-generation comprehensive VQA dataset consisting of the \textbf{trend-driven part} and the \textbf{static part}, containing $10,000$ videos across a wide spectrum of content types. The trend-driven part is based on the \textbf{TREND-Search} framework, which captures user preference profiles from trending lists on online platforms and formulates sampling strategies based on these profiles. The static part, on the other hand, is composed of supplementary samples selected from publicly available datasets. To support unified evaluation for various video types, we incorporate three evaluation dimensions: \textbf{technical}, \textbf{aesthetic}, and \textbf{AIGC-trace}. Experiments show that our dataset ensures high annotation quality and exhibits remarkable \textbf{generalization} across multiple content categories. In conclusion, our work presents a robust framework for advancing VQA, addressing challenges caused by the temporal evolution of user perceptual habits and preferences.
\end{abstract}

\section{Introduction}

Video media has evolved through several major \textbf{trends}, from real-world user-generated content~(UGC)~\cite{ghadiyaram2017capture}
and computer-generated~(CG) content~\cite{yu2022subjective} to professionally generated content~(PGC)~\cite{zheng2024sr4kvqa} with enhanced viewing quality, and subsequently to streaming-oriented media~\cite{duanmu2018quality} driven by the rise of online streaming platforms.  With the rapid advancement of platform-based short dramas and commercially deployed AI-generated content~(AIGC) videos, user preferences for video media trend have currently undergone a significant shift \cite{zhang2025large}. It is foreseeable that media trends are evolving rapidly. 
These trend shifts are profoundly reshaping the visual experiences of everyday users and offer valuable insights for the VQA field.
This raises a question: 
\textbf{\texttt{Should VQA datasets be continuously revised to align with these emerging trends?}}

The answer is affirmative. Current mainstream VQA datasets exhibit several limitations. Most datasets were developed before the prevalence of short dramas and AIGC videos~(detailed in \textit{supplementary materials~(Supp.)~Sec.~\ref{sec:Summary}}). They also focus on a single content type. For instance, UGC and AIGC videos are often evaluated separately. While such classifications hold value within specific content types, they cannot support \textbf{holistic} VQA across diverse content. Moreover, these datasets are temporally \textbf{static}, as they are largely extracted from open-source collections, without a robust update mechanism that allows the datasets to evolve in response to the rapidly changing media trends. These shortcomings underscore the critical need for the development of a next-generation, comprehensive VQA dataset.

To address these challenges (also presented in Fig.~\ref{fig:workflow}), our objective is to establish an \textbf{automated} and \textbf{dynamic} data collection framework, such that the resulting dataset can continuously adapt to both current and future media trends. This objective mainly involves three key points. First, \textbf{how to accurately characterize evolving media trends}. To this end, we model \textbf{user preferences from trending lists on online platforms}, since videos appearing on these lists naturally gain broader exposure, making them indicators of prevailing media trends. The second challenge lies in content/copyright safety. Furthermore, it is also crucial to design \textbf{evaluation dimensions} and \textbf{subjective annotation protocols} capable of coherently integrating videos from diverse content origins into a unified assessment framework. Therefore, we propose and develop \textbf{\textit{TREND-10K}}, a nex\textbf{\underline{t}}-generation comprehensive VQA dataset primarily based on user-p\textbf{\underline{r}}eference driv\textbf{\underline{en}} me\textbf{\underline{d}}ia content. Our contributions are threefold:

\begin{figure*}[t]
    \centering
    \includegraphics[width=0.95\linewidth]{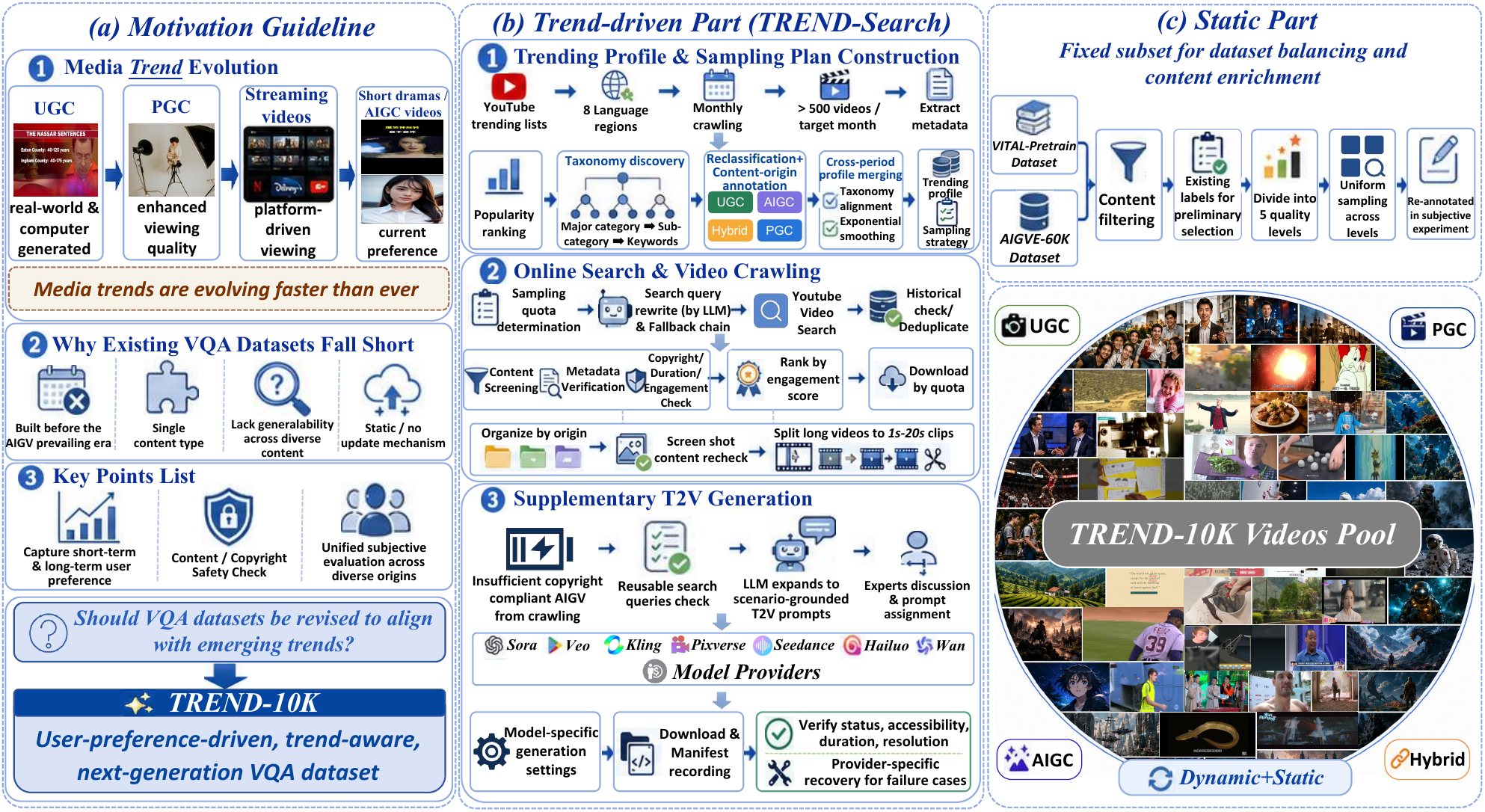}

    \caption{The motivation of \textit{TREND-10K} is to enable VQA to continuously adapt to evolving media-consumption trends, thereby better serving as feedback for media technologies. We design the \textit{TREND-Search} system to construct the trend-driven part through three major procedures, while introducing the static part for balancing. The resulting dataset contains videos with diverse content.}
    \label{fig:workflow}
\end{figure*}

\begin{itemize}
  \item First, we introduce the \textbf{TREND-Search} system, which automatically captures short-term user preference profiles from \textit{YouTube trending lists}, aggregates them into long-term user preference profiles, and effectively translates these preferences into sampling plans. It also facilitates the automatic online-crawling and content supervision, enabling periodic and dynamic updates to the database.
  \item Second, we construct the \textbf{TREND-10K} dataset, which consists of two components. First, the trend-based part contains videos collected over the past six months using \textit{TREND-Search} and supplemented with Text-to-Video~(T2V) videos generated from predefined sampling plans and expert discussions. The static part is selected from existing open-source datasets. We also develop a unified evaluation system encompassing three dimensions: \textit{technical}, \textit{aesthetic}, and \textit{AIGC-trace}. The dataset includes over $45K$ human-annotated labels.
  \item Finally, \textbf{extensive experiments} validate the intra-dataset consistency of \textit{TREND-10K} and its generalization ability, demonstrating its value for future VQA research.
\end{itemize}


\section{Related Works}
\subsection{UGC-VQA}
UGC-VQA mainly relies on subjective experiments, using mean opinion scores (MOS) as supervision for model development and benchmark evaluation. Existing datasets include \cite{nuutinen2016cvd2014,hosu2017konstanz,sinno2018large,wang2019youtube,ghadiyaram2017capture,ying2021patch,lu2024kvq,duan2025finevq,jia2025scaling} covering diverse scenarios and distortions and, in some cases, synthetically introduced distortions. 
Existing methods mainly differ in quality-aware representation learning. Knowledge-driven methods \cite{mittal2015completely,saad2014blind,tu2021ugc,tu2021rapique,duanmu2023bayesian} rely on hand-crafted natural scene statistics, distortion-sensitive descriptors, or handcrafted temporal features for quality prediction. More recent data-driven approaches~\cite{li2022blindly,sun2022deep,sun2024analysis,wang2021rich,wu2022fast,ying2021patch,zhang2023md,wen2024modular,wu2023exploring,qu2025kvq} leverage deep neural networks~(DNNs) to learn quality-relevant features, leading to improved prediction accuracy under diverse scenarios. With the rapid progress of large multimodal models~(LMMs), UGC-VQA methods built upon multimodal alignment improve robustness and generalization of VQA models. For instance, \cite{wu2024q1,jia2024vqa,jia2025vital} demonstrate strong performance in UGC-VQA scoring by exploiting quality-aware multimodal representations.

\subsection{T2V AIGC-VQA}

Recent advances in T2V generation have inspired various evaluation benchmarks and metrics. The AIGC-VQA datasets covers preference comparison~\cite{wang2024aigv}, multi-dimensional MOS prediction~\cite{chivileva2023measuring,zhang2024benchmarking,wang2025love}, and assessment of specified attributes~\cite {chen2024gaia,cao2025agav}. Fine-MOS datasets provide finer subjective supervision. Metrics such as \cite{guan2025etva} and \cite{li2024evaluating} improve text-video correspondence evaluation. LMMs~\cite{wang2024aigv,wang2025love,cao2025agav,wang2025tdve} also exhibit strong visual question-answering ability. \cite{huang2024vbench} further introduces multiple task-specific detection models.

However, existing UGC and AIGC VQA datasets are mostly constructed before the large-scale commercialization of short dramas and AIGC videos. As a result, UGC-VQA datasets are primarily centered on real-shot and screen-captured recordings, while videos in existing AIGC-VQA datasets are often extremely short and generated by relatively early-stage models with evident AIGC artifacts and abnormalities, which substantially limits their value for realistic quality evaluation. Existing UGC and AIGC datasets follow different evaluation protocols, preventing unified assessment. Consequently, models trained on them struggle to generalize to emerging media content. This limitation motivates \textit{TREND-10K}.

\section{The TREND-10K}

Constructing \textit{TREND-10K} faces three challenges. First, user preferences are often diverse and rapidly evolving, requiring a structured representation of contemporary media content. Second, copy-right and safety constraints prevent direct use of trending videos. We instead retrieve semantically relevant alternatives from copyright-cleared and safety-compliant sources. Third, to further enrich the AIGC component of the dataset, we use the extracted trends as T2V prompts and synthesize videos with multiple AIGC models. Together, they form the \textit{trend-driven} part~(with $7,000$ videos) of \textit{TREND-10K} and the \textit{TREND-Search} pipeline. Finally, to improve quality, diversity, and annotation balance, we re-annotate $3,000$ videos from open-source datasets. Statistics of the dataset composition are presented in \textit{Supp.~Sec.~\ref{sec:dataset_composition}}.  This portion of the dataset constitutes the \textit{static part}. The detailed construction processes are shown in Fig.~\ref{fig:workflow} and detailed in \textit{Supp.~Sec.~\ref{sec:D1},~\ref{sec:D2}, and~\ref{sec:D3}}.

\subsection{Trend-driven Part Construction}
\paragraph{Trending Profile and Sampling Plan Construction}
\textit{TREND-Search} automatically collects videos from \textit{YouTube Trending List} from \textit{eight language regions}~(\textit{US, GB, JP, KR, BR, IN, DE, FR}) following a \textit{monthly crawling} protocol. For each target month, it retrieves \textit{more than $1000$ listed videos} and retains their metadata, together with titles, descriptions, tags, category IDs, and popularity statistics. The collected videos are then \textit{ranked} by popularity.
Using metadata, the system constructs a \textit{taxonomy} through dynamic category discovery. Specifically, it assigns each video to a major category according to its category ID and groups videos by category. The grouped metadata are then processed by an LLM~(we use \textit{GPT-5-Nano}~\cite{singh2025openai} here) to induce subcategories. Within each major category, the LLM identifies fine-grained subcategories from video titles, descriptions, and tags, and further summarizes $5$--$10$ representative keywords for each subcategory. The resulting major-category--subcategory--keyword hierarchy forms the taxonomy.

The taxonomy is then used to \textit{reclassify} all collected videos. Within each major category, each video is assigned to the most semantically appropriate taxonomy based on its title, description, and tags. Here, we set the final classification result to the major-category--subcategory levels for reliable retrieval in subsequent online crawling steps. Concurrently, the LLM annotates the content origin of each video based on its metadata, categorizing it as \textit{UGC}, \textit{AIGC}, \textit{Hybrid}, or \textit{PGC}. This process results in a joint distribution over content origin, major category, and subcategory for the trending videos in the given period.
To integrate trends across multiple periods, it employs a dynamic profile merging strategy. It first \textit{aligns} major categories and subcategories across consecutive periods to ensure semantic consistency. It updates their contributions to the global trending profile using \textit{exponential smoothing}. Finally, it aggregates all taxonomy distributions into a unified \textit{trending profile} and converts it into an executable \textit{sampling strategy}.

\paragraph{Online Search and Video Crawling}
\textit{TREND-Search} performs online search to execute the sampling plan and prepare dataset-ready video segments. The system first retains only those items in the sampling plan where the target count is greater than $10$, thus defining the \textit{planned quota} for the online search. To avoid duplicates, the system then \textit{checks historical sampling records}, extracts the \textit{video id} values, and excludes any previously collected videos from subsequent download steps. To ensure that the original sampling intent is preserved while facilitating retrieval, each sampling plan item retains its original search query and defines a fallback chain for the search. The fallback chain progressively relaxes the query from its original taxonomy. Consequently, for each search query, a corresponding suite of fallback queries is derived to handle cases where no valid video is found. The system further uses the LLM to rewrite the search query to resemble a human-authored one before using it for search. 

Videos are first retrieved via direct \textit{YouTube} search, followed by metadata validation. For each query, the system sequentially applies fallback queries and stops once enough candidates are retrieved. To eliminate invalid videos, the system first makes a coarse \textit{ content
screening } using LLM at the candidate search stage, skipping any video that is inappropriate or irrelevant.
Fine-grained filtering is applied after candidate retrieval through a thorough \textit{metadata verification} process. At this stage, candidates are filtered based on copyright, duration, engagement metrics~(views, likes, and comments), and metadata completeness. After filtering, the remaining candidates are \textit{sorted by their engagement scores}, and the system selects items for \textit{download} based on the target quota for each query.
The downloaded files are \textit{organized by content origins}. To further exclude unsuitable videos, the system performs a \textit{screenshot-based recheck}. The collected long videos are then randomly \textit{split into short clips}, retaining non-overlapping valid clips within the valid duration range~($1$s-$20$s). 

\paragraph{Supplementary T2V Generation}
Given the \textit{limited availability of copyright-compliant AIGC videos} from online crawling, the system further augments the preceding sampling plan by constructing T2V prompts and using them as inputs to state-of-the-art~(\textit{SOTA}) AIGC T2V models. Specifically, the system reuses the search queries and calls the LLM to first determine whether each query is \textit{suitable for T2V prompt rewriting}. For each approved query, the LLM further expands it into a scenario-grounded T2V prompt with at least $50$ words, providing richer visual and semantic guidance for T2V generation. In addition, we organize an \textit{expert discussion} regarding the limitations of current commercial T2V models to design another supplementary set of T2V prompts. These prompts place greater emphasis on challenging generation scenarios, including dynamic motions, intricate object interactions, and professional cinematographic perspectives. We then generate videos using multiple \textit{model providers}, including \textit{Sora}, \textit{VEO}, \textit{Kling}, \textit{PixVerse}, \textit{Seedance}, \textit{Hailuo}, and \textit{Wan}~(model details are in \textit{Supp. Sec.~\ref{sec:T2Vmodels}}). Under this practical workflow, the \textit{TREND-Search} first selects prompts from the constructed pool and then randomly assigns each prompt to a model provider. Each \textit{provider-specific generation pipeline} controls the generation configuration, such as resolution, duration, and aspect ratio. Once a generation request succeeds, the resulting video is immediately downloaded and recorded in a \textit{manifest JSON file}. This workflow further incorporates a lightweight \textit{verification stage} for generated videos. Specifically, the system validates each output by checking the response status and the absence of recorded errors in the manifest JSON. For failed or incomplete generations, provider-specific \textit{recovery strategies} are adopted.

Except for a very limited amount of expert intervention, the above three steps can be completed in an almost fully automated manner. This enables the \textit{TREND-Search} system to dynamically acquire trend-aligned video content through scheduled and continuous collection procedures.

\subsection{Static Part Construction}
To balance the dataset and ensure sufficient coverage of both real-world and synthetic content, we introduce the \textit{TREND-10K} static part, a fixed subset independent of the \textit{TREND-Search} process. This subset is sampled from the open-source \textit{VITAL-Pretrain}~\cite{jia2025vital} and \textit{AIGVE-60K}~\cite{wang2025love} datasets. After necessary \textit{content filtering}, we use the existing visual quality labels in these datasets for preliminary selection. Following~\cite{wu2024q1}, videos within each dataset are \textit{divided into $5$ quality levels} according to their dataset-specific score distributions, and samples are uniformly drawn across all levels. All selected videos are subsequently \textit{re-annotated} through our subjective experiments to ensure higher annotation reliability and consistency.

\subsection{Subjective Experiment}

After collecting all videos, we conduct a carefully designed subjective experiment to obtain representative, multidimensional human annotations for holistic quality assessment across diverse video content origins. To this end, we define three evaluation dimensions.
The first dimension is \textbf{technical quality}, which evaluates both visual fidelity and temporal consistency. It includes factors such as \textit{perceived clarity}, \textit{noise}, \textit{compression artifacts}, \textit{exposure}, \textit{color}, \textit{playback smoothness}, \textit{frame stability}, and \textit{temporal inconsistencies in AIGC content}.
The second dimension is \textbf{aesthetic quality}, which measures the aesthetic expressiveness of a video beyond basic technical fidelity. It covers attributes such as \textit{composition}, \textit{lighting}, \textit{organization}, \textit{rhythm}, \textit{emotional atmosphere}, and \textit{cinematic expressiveness}, assessing whether the visual presentation is aesthetically appealing and semantically coherent.
The third dimension is \textbf{perceptual authenticity}, namely \textbf{the degree of perceived AIGC-trace}. Rather than relying on explicit semantic cues such as watermarks or titles, this dimension evaluates \textit{how strongly AI-generated characteristics can be perceived directly from the video content itself}.
\begin{figure*}[t]
    \centering
    \includegraphics[width=\linewidth]{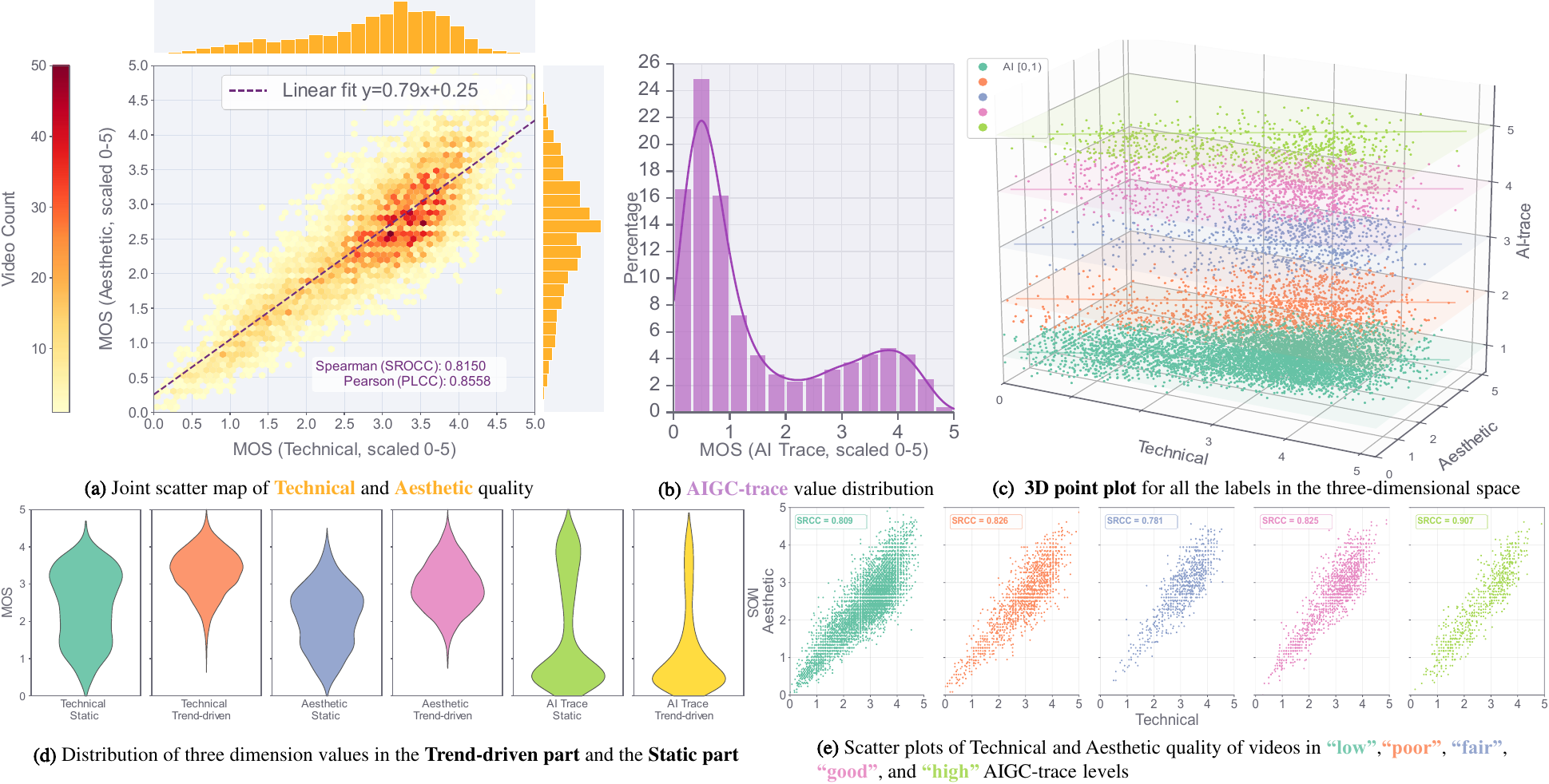}
    \caption{Visualization of annotated data analysis.}
    \label{fig:data_analysis}
\end{figure*}

The rationale behind this three-dimensional design lies in the complementary roles of the three dimensions. Technical and aesthetic quality serve as the fundamental anchors of most media content, since they jointly determine the \textbf{basic watchability} of a video across diverse content sources. Technical quality reflects the degree of visual fidelity and completion, while aesthetic quality measures the viewer's affinity and expressive appeal once the content is perceptually understandable. 
Perceptual authenticity further unifies videos generated by different production mechanisms within a common perceptual framework. It characterizes the similarity between the viewed video and the user's internal perceptual anchor of ``human-generated authentic'' media content. Since this perceptual anchor is shaped through long-term media consumption, it is largely shared among users with similar cultural and educational backgrounds.

We recruit over $180$ college students with abundant background knowledge, all sharing the same nationality and cultural background, with an age range of $20$-$25$. Each dimension is rated using $1$-$5$ \textit{Likert scale}. To facilitate management and ensure data privacy, we develop a unified \textit{online annotation platform}. 
The dataset is randomly divided into $10$ groups, each containing $1{,}000$ videos with an equal proportion from the two parts. Following strict predefined guidelines~(detailed in \textit{Supp.~Sec.\ref{sec:guideline}}), subjects annotate their assigned video groups on monitors with resolutions of at least $1080$p, while also performing content supervision to ensure compliance with the required standards. Videos that fail to satisfy the supervision criteria are removed and replaced with backup videos from the same part.
To ensure scoring consistency, each annotator first \textit{undergoes complete training}, including tutorial reading and practice on $32$ training videos. Each training video is accompanied by reference scores and explanatory notes for all dimensions. After annotation begins, we further introduce a \textbf{warm-up mechanism}: the first $100$ videos annotated by each subject are excluded from the final results and used solely for practice, while the subjects remain unaware of this setting during the experiment. This strategy helps annotators establish stable and consistent scoring criteria.
After annotation is completed, we manually inspect the score distributions and annotation durations of each annotator across all three dimensions. We exclude annotations from subjects whose score distributions appear as outliers or whose annotation durations exhibit abnormal patterns, such as unusually rapid completion. Following the recommendations of \textbf{ITU-R BT.500}~\cite{series2012methodology}, each video finally receives annotations from at least $15$ distinct annotators. More details are in \textit{Supp.~Sec.~\ref{sec:Annotation}}.

To obtain the final scores consistently, we first compute the average score of all annotators for each video and each evaluation dimension. We then uniformly linearly scale the remaining scores to the range from $0$ to $5$. In addition, for each video, annotation-score outliers that deviate from the mean by more than three standard deviations are excluded before the final averaging process.
Finally, we divide the quality range into five levels with an interval of one point, resulting in the labels ``\texttt{low}'' for $0$--$1$, ``\texttt{poor}'' for $1$--$2$, ``\texttt{fair}'' for $2$--$3$, ``\texttt{good}'' for $3$--$4$, and ``\texttt{high}'' for $4$--$5$.

\section{Data Analysis and Experiments}

After constructing the \textit{TREND-10K} and obtaining all subjective labels, we focus on three main questions. \ding{182} How are the annotations distributed across the three dimensions, and what correlations or additional patterns can be observed? \ding{183} How effectively do representative SOTA VQA methods adapt to \textit{TREND-10K} under intra-dataset and cross-dataset evaluation settings? \ding{184} As a next-generation comprehensive VQA dataset, to what extent do models trained on \textit{TREND-10K} generalize to other widely used content-specific VQA datasets? We discuss these questions in the following sections.

 \begin{table}[H]      
\centering      
\tiny      
\renewcommand\arraystretch{1.2}      
\renewcommand\tabcolsep{1.8pt}                
\resizebox{\linewidth}{!}{%
\rowcolors{2}{gray!12}{white}
\begin{tabular}{c|cccc}      
\hline      
\multicolumn{1}{c|}{\textbf{Models}} & \textbf{Technical} & \textbf{Aesthetic} & \textbf{AIGC-trace} & \textbf{AVG.} \\      
\hline      
\textsc{FAST-VQA-B}     
& 0.833~/~0.856      
& 0.801~/~0.818      
& 0.846~/~0.858      
& 0.835 \\
      \textsc{DOVER}    
& 0.838~/~0.851      
& \textit{0.821}~/~\textit{0.833}     
& 0.858~/~0.907      
& 0.851 \\
      \textsc{Mini-VQA-VII}
& 0.820~/~0.844      
& 0.796~/~0.803      
& 0.851~/~0.897      
& 0.835 \\
      \textsc{Mini-VQA-IX}      
& 0.825~/~0.867      
& 0.813~/~0.821      
& 0.845~/~0.891      
& 0.844 \\
      \textsc{KVQ}   
& \textit{0.850}~/~\textit{0.879}      
& 0.805~/~0.817      
& \textbf{0.880}~/~\textbf{0.913}      
& \textit{0.857} \\
      \textsc{Q-Align} 
& \textbf{0.856}~/~0.877      
& 0.812~/~0.821      
& \textit{0.866}~/~0.905      
& 0.856 \\
      \textsc{VITAL}  
& 0.844~/~\textbf{0.884}     
& \textbf{0.826}~/~\textbf{0.837}  
& 0.857~/~\textit{0.910}    
& \textbf{0.860} \\      
\hline      
\end{tabular}}  
\caption{Intra-dataset results~(\textit{PLCC}$\uparrow$~/~\textit{SRCC}$\uparrow$). \textit{Mini-VQA} refers to \textit{Minimalistic-VQA}. [Per column: best in  
\textbf{boldface}, second in \textbf{italics}.]}
\label{tab:intra}  
\end{table}
\begin{table*}[!htb]
        \centering
        \renewcommand\arraystretch{1.05}
        \renewcommand\tabcolsep{0.9pt}
        \belowrulesep=0pt\aboverulesep=0pt

        \resizebox{\textwidth}{!}{%
        \begin{tabular}{l|ccccccccccccccccc}
        \hline
        \multicolumn{1}{l|}{\textbf{Datasets}}
          & \multicolumn{2}{c}{\textbf{YT-UGC}}
          & \multicolumn{2}{c}{\textbf{\textit{YT-Gaming}}}
          & \multicolumn{2}{c}{\textbf{\textit{CGVDS}}}
          & \multicolumn{2}{c}{\textbf{\textit{KVQ}}}
          & \multicolumn{2}{c}{\textbf{\textit{FINEVD}}}
          & \multicolumn{2}{c}{\textbf{\textit{AIGVQA}}}
          & \multicolumn{2}{c}{\textbf{\textit{AGAV}}}
          & \multicolumn{2}{c}{\textbf{\textit{GAIA}}}
          & \multirow{3}{*}{\textbf{AVG.}}\\
        \cline{1-17}
        \multicolumn{1}{l|}{\textbf{\# test videos}}
           & \multicolumn{2}{c}{\textbf{1,098}}
          & \multicolumn{2}{c}{\textbf{600}}
          & \multicolumn{2}{c}{\textbf{357}}
          & \multicolumn{2}{c}{\textbf{2,926}}
          & \multicolumn{2}{c}{\textbf{1,016}}
          & \multicolumn{2}{c}{\textbf{2,808}}
          & \multicolumn{2}{c}{\textbf{622}}
          & \multicolumn{2}{c}{\textbf{9,180}}
          & \\
        \cline{1-17}
          \textbf{Models}
          & SRCC$\uparrow$ & PLCC$\uparrow$
          & SRCC$\uparrow$ & PLCC$\uparrow$
          & SRCC$\uparrow$ & PLCC$\uparrow$
          & SRCC$\uparrow$ & PLCC$\uparrow$
          & SRCC$\uparrow$ & PLCC$\uparrow$
          & SRCC$\uparrow$ & PLCC$\uparrow$
          & SRCC$\uparrow$ & PLCC$\uparrow$
          & SRCC$\uparrow$ & PLCC$\uparrow$
          & \\
        \cdashline{1-18}
          \multicolumn{18}{l}{\textit{Trained on LSVQ~(train)~($28,056$ videos)}}\\
          \cdashline{1-18}
        \rowcolor{light-gray0} \textsc{FAST-VQA-B}
          & 0.730 & 0.743 & 0.636 & 0.675 & 0.727 & 0.750 & 0.524 & 0.529 & 0.617 & 0.669 & 0.533 & 0.589 & 0.315 & 0.315 & 0.228 & 0.228 &
  0.550 \\
        \textsc{Mini-VQA-VII}
          & 0.748 & 0.756 & 0.621 & 0.693 & 0.725 & 0.769 & 0.612 & 0.654 & 0.550 & 0.632 & 0.625 & 0.634 & 0.277 & 0.281 & 0.207 & 0.199 &
  0.561 \\
         \rowcolor{light-gray0} \textsc{Mini-VQA-IX}
          & 0.761 & 0.768 & 0.675 & 0.734 & 0.767 & 0.806 & 0.629 & 0.668 & 0.625 & 0.682 & 0.643 & 0.644 & 0.268 & 0.279 & 0.175 & 0.175 &
  0.581 \\
        \textsc{DOVER}
          & 0.725 & 0.711 & 0.594 & 0.627 & 0.702 & 0.724 & 0.527 & 0.547 & 0.552 & 0.576 & 0.698 & 0.696 & 0.298 & 0.311 & 0.213 & 0.212 &
  0.545 \\
        \rowcolor{light-gray0} \textsc{KVQ}
          & 0.791 & 0.816 & 0.647 & 0.700 & 0.743 & 0.783 & 0.533 & 0.550 & 0.545 & 0.631 & 0.384 & 0.373 & 0.357 & 0.354 & 0.240 & 0.238 &
  0.543 \\
        \textsc{Q-Align}
          & 0.836 & 0.849 & 0.611 & 0.682 & 0.757 & 0.800 & 0.613 & 0.655 & 0.596 & 0.676 & 0.723 & 0.715 & 0.320 & 0.310 & 0.257 & 0.235 &
  0.602 \\
        \rowcolor{light-gray0} \textsc{VITAL}
          & \underline{0.846} & \underline{0.849} & 0.710 & 0.764 & 0.764 & 0.804 & 0.644 & 0.666 & 0.647 & 0.612 & 0.687 & 0.713 & 0.328 & 0.235
  & 0.221 & 0.196 & 0.605 \\
        \cdashline{1-18}
          \multicolumn{18}{l}{\textit{Trained on TREND-10K~(technical)}}\\
          \cdashline{1-18}
        \textsc{FAST-VQA-B}
          & 0.758 & 0.768 & 0.684 & 0.728 & 0.746 & 0.770 & 0.590 & 0.594 & 0.682 & 0.716 & 0.603 & 0.630 & 0.396 & 0.405 & 0.296 & 0.302 &
  0.604 \\
        \rowcolor{light-gray0} \textsc{Mini-VQA-VII}
          & 0.774 & 0.780 & \underline{0.713} & 0.766 & 0.739 & 0.740 & \underline{0.671} & 0.645 & \textit{0.746} & 0.722 &
  0.674 & 0.666 & 0.353 & 0.383 & 0.331 & 0.332 & 0.627 \\
        \textsc{Mini-VQA-IX}
          & 0.795 & 0.802 & \textbf{0.741} & \textbf{0.791} & 0.766 & 0.788 & 0.632 & 0.643 &
  \underline{0.725} & \textit{0.770} & 0.715 & 0.712 & 0.355 & 0.373 & \textit{0.357} & 0.330 & 0.643 \\
        \rowcolor{light-gray0} \textsc{DOVER}
          & 0.787 & 0.787 & 0.701 & \underline{0.771} & \underline{0.776} & 0.804 & 0.629 & 0.642 & 0.719 & 0.742 & \underline{0.754} &
  \underline{0.738} & \textbf{0.439} & \textbf{0.457} & 0.313 & 0.325 & \underline{0.649} \\
        \textsc{KVQ}
          & 0.801 & 0.813 & 0.691 & 0.745 & 0.762 & 0.797 & 0.612 & 0.586 & 0.690 & \underline{0.746} & 0.442 & 0.438 &
  \textit{0.438} & \underline{0.441} & 0.288 & 0.274 & 0.598 \\
        \rowcolor{light-gray0} \textsc{Q-Align}
          & \textit{0.848} & \textbf{0.857} & 0.668 & 0.735 & \textit{0.779} &
  \textbf{0.818} & 0.645 & \textit{0.688} & 0.662 & 0.724 & 0.748 & \textit{0.742} &
  0.420 & 0.412 & 0.308 & \underline{0.332} & \textit{0.649} \\
        \textsc{VITAL}
          & \textbf{0.851} & 0.831 & \textit{0.733} & \textit{0.772} &
  \textbf{0.794} & \textit{0.816} & \textit{0.678} & \underline{0.681} &
  \textbf{0.765} & \textbf{0.781} & \textit{0.767} & 0.734 & \underline{0.434} &
  \textit{0.447} & \underline{0.350} & \textbf{0.367} & \textbf{0.675} \\
        \cdashline{1-18}
          \multicolumn{18}{l}{\textit{Trained on TREND-10K~(aesthetic)}}\\
          \cdashline{1-18}
       \rowcolor{light-gray0}  \textsc{FAST-VQA-B}
          & 0.746 & 0.758 & 0.670 & 0.712 & 0.738 & 0.762 & 0.568 & 0.574 & 0.660 & 0.704 & 0.586 & 0.614 & 0.335 & 0.342 & 0.274 & 0.281 &
  0.583 \\
        \textsc{Mini-VQA-VII}
          & 0.770 & 0.784 & 0.700 & 0.752 & 0.738 & 0.776 & 0.655 & 0.661 & 0.688 & 0.704 & 0.650 & 0.648 & 0.338 & 0.344 & 0.282 & 0.276 &
  0.610 \\
      \rowcolor{light-gray0}   \textsc{Mini-VQA-IX}
          & 0.822 & 0.824 & 0.704 & 0.754 & 0.775 & \underline{0.812} & 0.646 & 0.675 & 0.696 & 0.731 & 0.690 & 0.681 & 0.332 & 0.343 & 0.286 &
  0.270 & 0.628 \\
        \textsc{DOVER}
          & 0.762 & 0.760 & 0.673 & 0.733 & 0.748 & 0.775 & 0.588 & 0.604 & 0.681 & 0.712 & 0.728 & 0.716 & 0.366 & 0.382 & 0.286 & 0.292 &
  0.613 \\
       \rowcolor{light-gray0}  \textsc{KVQ}
          & 0.798 & 0.822 & 0.668 & 0.724 & 0.754 & 0.789 & 0.588 & 0.575 & 0.666 & 0.706 & 0.420 & 0.405 & 0.394 & 0.392 & 0.262 & 0.258 &
  0.576 \\
        \textsc{Q-Align}
          & 0.840 & \textit{0.852} & 0.642 & 0.704 & 0.766 & 0.807 & 0.625 & 0.671 & 0.638 & 0.697 & 0.732 & 0.726 & 0.332
  & 0.326 & 0.270 & 0.252 & 0.618 \\
     \rowcolor{light-gray0}    \textsc{VITAL}
          & 0.767 & 0.775 & 0.629 & 0.675 & 0.735 & 0.750 & \textbf{0.715} & \textbf{0.718} & 0.713 &
  0.699 & \textbf{0.785} & \textbf{0.779} & 0.402 & 0.402 & \textbf{0.381} &
  \textit{0.342} & 0.642 \\
        \cline{1-18}
        \end{tabular}%
        }
        \caption{Cross-dataset performance of VQA models pre-trained \textit{from scratch} on different sources. Test
  datasets marked in \textit{italics} denote \textit{out of distribution~(OOD)}. When trained on \textit{LSVQ}, all models use publicly available weights, except \textit{VITAL}, which uses its dedicatedly pretrained \textit{VITAL-Base}. [Per column: highest in \textbf{boldface}, second in
  \textbf{italics}, third \textbf{underlined}.]}
        \label{tab:pretrain_inter}
    \end{table*}

\subsection{Data Analysis}
We evaluate \textit{inter-annotator agreement}~(IAA) for the three dimensions using \textit{Krippendorff's Alpha}~\cite{krippendorff1970estimating}. The values are $0.621$, $0.586$, and $0.677$ for technical, aesthetic, and AIGC-trace scores, respectively, indicating reliable agreement while retaining individual variability. We further visualize the joint distribution of technical and aesthetic scores with their marginal histograms~(shown in Fig.~\ref{fig:data_analysis}~(a)). Both dimensions are approximately \textit{Gaussian}-distributed~(with technical quality showing a mild right skew) and most videos fall within the score ranges of $2$--$3$ and $3$--$4$, consistent with established VQA datasets like \textit{LSVQ}~\cite{ying2021patch}. The two dimensions show a strong positive correlation, with an SRCC of $0.815$. Together with the relatively lower \textit{IAA} of aesthetic quality, this suggests that \textbf{aesthetic quality can be regarded as a higher-level manifestation of technical quality}: technical fidelity strongly influences aesthetic perception. In contrast, aesthetic judgments remain more subjective due to personal preference and cultural background.

We also examine the distribution of AIGC-trace scores~(Fig.~\ref{fig:data_analysis}~(b)), the three-dimensional $3$-D scatter plot~(Fig.~\ref{fig:data_analysis}~(c)), and the technical-aesthetic correlations within each of the $5$ AIGC-trace levels~(Fig.~\ref{fig:data_analysis}~(e)). Most videos are concentrated in the low AIGC-trace range~($0$--$1$). Meanwhile, technical and aesthetic scores remain positively correlated across all AIGC-trace levels, indicating that their strong coupling extends from conventional UGC videos to broader content sources and supporting the validity of our three-dimensional evaluation design. The highest AIGC-trace level~($4$--$5$) shows the strongest technical-aesthetic correlation, with an SRCC of $0.907$, suggesting that for videos with evident AI-generated traces, basic visual fidelity almost directly determines their aesthetic appeal. Finally, the violin plots in Fig.~\ref{fig:data_analysis}~(d) compare the static and trend-driven components of \textit{TREND-10K}. The trend-driven portion exhibits noticeably higher technical and aesthetic quality distributions, reflecting the evolving quality characteristics of contemporary online media content. In contrast, the static component plays an important role in regularizing and balancing the overall dataset distribution across different quality levels.

  \begin{table*}
  \small
    \centering
    \renewcommand\arraystretch{1.2}
    \renewcommand\tabcolsep{0.4pt}
    \belowrulesep=0pt\aboverulesep=0pt
    \resizebox{\textwidth}{!}{%
    \begin{tabular}{l|cccccccccccccccccc}
    \hline
    \multicolumn{1}{l|}{\textbf{Datasets}}
      & \multicolumn{2}{c}{\textbf{LSVQ~(test)}}
      & \multicolumn{2}{c}{\textbf{LSVQ~(1080p)}}
      & \multicolumn{2}{c}{\textbf{KoNViD-1K}}
      & \multicolumn{2}{c}{\textbf{YT-UGC}}
      & \multicolumn{2}{c}{\textbf{\textit{YT-Gaming}}}
      & \multicolumn{2}{c}{\textbf{\textit{KVQ}}}
      & \multicolumn{2}{c}{\textbf{\textit{FINEVD}}}
      & \multicolumn{2}{c}{\textbf{\textit{AIGVQADB}}}
      & \multicolumn{2}{c}{\textbf{\textit{GAIA}}}\\
    \cline{1-19}
    \multicolumn{1}{l|}{\textbf{\# test videos}}
      & \multicolumn{2}{c}{\textbf{7,182}}
      & \multicolumn{2}{c}{\textbf{3,573}}
      & \multicolumn{2}{c}{\textbf{1,200}}
      & \multicolumn{2}{c}{\textbf{1,098}}
      & \multicolumn{2}{c}{\textbf{600}}
      & \multicolumn{2}{c}{\textbf{2,926}}
      & \multicolumn{2}{c}{\textbf{1,016}}
      & \multicolumn{2}{c}{\textbf{2,808}}
      & \multicolumn{2}{c}{\textbf{9,180}}\\
    \cline{1-19}
    \textbf{Models}
      & SRCC$\uparrow$ & PLCC$\uparrow$
      & SRCC$\uparrow$ & PLCC$\uparrow$
      & SRCC$\uparrow$ & PLCC$\uparrow$
      & SRCC$\uparrow$ & PLCC$\uparrow$
      & SRCC$\uparrow$ & PLCC$\uparrow$
      & SRCC$\uparrow$ & PLCC$\uparrow$
      & SRCC$\uparrow$ & PLCC$\uparrow$
      & SRCC$\uparrow$ & PLCC$\uparrow$
      & SRCC$\uparrow$ & PLCC$\uparrow$\\
    \cdashline{1-19}
    \multicolumn{19}{l}{\textit{PreTrained on LSVQ~(train)+Finetuned on KVQ}}\\
    \cdashline{1-19}
    \textsc{FAST-VQA-B}
      & 0.689 & 0.688 & 0.532 & 0.557 & 0.542 & 0.556 & 0.657 & 0.662 & 0.678 & 0.703 & - & - & 0.698 & 0.719 & 0.296 & 0.378 & 0.110 & 0.107\\
   \rowcolor{light-gray0}  \textsc{DOVER}
      & 0.789 & 0.771 & 0.677 & 0.703 & 0.663 & 0.667 & 0.774 & 0.773 & 0.686 & 0.736 & - & - & 0.704 & 0.735 & 0.317 & 0.319 & 0.188 & 0.185\\
    \textsc{KVQ}
      & 0.764 & 0.762 & 0.662 & 0.716 & 0.702 & 0.720 & 0.769 & 0.765 & 0.714 & 0.732 & - & - & 0.734 & 0.756 & 0.323 & 0.323 & 0.210 & 0.195\\
   \rowcolor{light-gray0}  \textsc{VITAL}
      & 0.797 & 0.800 & 0.658 & 0.720 & 0.717 & 0.762 & 0.758 & 0.782 & 0.732 & 0.771 & - & - & 0.741 & 0.737 & 0.358 & 0.367 & 0.181 & 0.201\\
    \cdashline{1-19}
    \multicolumn{19}{l}{\textit{PreTrained on LSVQ~(train)+Finetuned on AIGVQADB}}\\
    \cdashline{1-19}
    \textsc{FAST-VQA-B}
      & 0.699 & 0.709 & 0.620 & 0.652 & 0.579 & 0.607 & 0.633 & 0.639 & 0.548 & 0.588 & 0.230 & 0.235 & 0.393 & 0.425 & - & - & \textit{0.362} & \textit
  {0.364}\\
   \rowcolor{light-gray0}  \textsc{DOVER}
      & 0.743 & 0.738 & 0.707 & 0.744 & 0.725 & 0.747 & 0.735 & 0.742 & 0.607 & 0.672 & 0.468 & 0.488 & 0.474 & 0.531 & - & - & 0.290 & 0.292\\
    \textsc{KVQ}
      & 0.722 & 0.728 & 0.590 & 0.652 & 0.637 & 0.659 & 0.688 & 0.705 & 0.544 & 0.599 & 0.398 & 0.430 & 0.371 & 0.432 & - & - & 0.294 & 0.297\\
   \rowcolor{light-gray0}  \textsc{VITAL}
      & 0.771 & 0.756 & 0.685 & 0.711 & 0.732 & 0.741 & 0.721 & 0.752 & 0.647 & 0.702 & 0.312 & 0.374 & 0.672 & 0.651 & - & - & 0.331 & 0.305\\
    \cdashline{1-19}
    \multicolumn{19}{l}{\textit{PreTrained on LSVQ~(train)+Finetuned on TREND10K-technical}}\\
    \cdashline{1-19}
    \textsc{FAST-VQA-B}
      & 0.787 & 0.789 & 0.692 & 0.761 & 0.787 & 0.796 & 0.770 & 0.770 & \textit{0.766} & \underline{0.804} & \textit{0.694} & \textit{0.685} &
  \underline{0.757} & \underline{0.759} & 0.724 & \underline{0.745} & 0.342 & 0.339\\
   \rowcolor{light-gray0}  \textsc{DOVER}
      & \textit{0.816} & \textit{0.812} & \underline{0.739} & \underline{0.770} & \textit{0.805} & \underline{0.811} & \textit{0.802} &
  \underline{0.805} & \underline{0.746} & \textit{0.811} & \underline{0.614} & 0.621 & 0.741 & \textit{0.784} & \textit{0.751} & 0.736 & \textbf{0.352} &
  \textbf{0.354}\\
    \textsc{KVQ}
      & \underline{0.809} & \underline{0.810} & \textbf{0.767} & \textit{0.803} & \underline{0.799} & \textit{0.814} & \underline{0.801} & \textit{0.813}
  & 0.691 & 0.745 & 0.609 & \underline{0.632} & \textbf{0.788} & 0.754 & \underline{0.733} & \textit{0.756} & 0.293 & 0.297\\
   \rowcolor{light-gray0}  \textsc{VITAL}
      & \textbf{0.843} & \textbf{0.838} & \textit{0.757} & \textbf{0.822} & \textbf{0.831} & \textbf{0.835} & \textbf
  {0.851} & \textbf{0.848} & \textbf{0.778} & \textbf{0.829} & \textbf{0.722} & \textbf{0.725} & \textit{0.769} &
  \textbf{0.791} & \textbf{0.785} & \textbf{0.773} & \textbf{0.402} & \textbf{0.386}\\
    \cline{1-19}
    \end{tabular}%
    }
    \caption{Transfer learning performance of VQA models trained on top of \textit{LSVQ~(train)} on different source datasets and evaluated on multiple test datasets.
    \textbf{Intra-dataset evaluation is not conducted.}}
    \label{tab:transfer_inter}
  \end{table*}

\subsection{Experiments}
\paragraph{Intra-dataset Evaluation}
For intra-dataset evaluation, we select the most representative \textit{SOTA} VQA models, including DNN-based models \textit{FAST-VQA}~\cite{wu2022fast}, \textit{DOVER}~\cite{wu2023exploring}, \textit{Minimalistic-VQA}~\cite{sun2024analysis}, and \textit{KVQ}~\cite{qu2025kvq}, as well as LMM-based models \textit{Q-Align}~\cite{wu2024q1}, and \textit{VITAL}~\cite{jia2025vital}. We report both the \textit{Pearson linear correlation coefficient~(PLCC)} and the \textit{Spearman rank-order correlation coefficient~(SRCC)}, which are standard metrics for VQA model evaluation. For \textit{VITAL}, since each video in \textit{TREND-10K} is annotated by at least $15$ subjects, we can construct a valid opinion distribution which satisfies the training requirement.  
For DNN-based methods, we split \textit{TREND-10K} into training, validation, and test sets with a ratio of $7\!\!:\!\!1\!\!:\!\!2$. 
For LMM-based methods, because their original settings do not rely on a validation set, we use the same random split but discard the validation set. We conduct a $5$-fold train-evaluation, training each dimension independently, and report the average results over the $
5$ runs. The results are reported in Tab.~\ref{tab:intra}.

\paragraph{Cross-dataset Generalization Evaluation}
We evaluate cross-dataset generalization from two perspectives. First, we treat \textit{TREND-10K} as the direct pre-training source~(use \textit{from scratch} training strategy on all models) and compare it with the most commonly used VQA training dataset \textit{LSVQ~(train)}~\cite{ying2021patch} on a diverse set of content-specific benchmarks. These benchmarks include UGC mixed content datasets \textit{LIVE-YT-UGC}~\cite{wang2019youtube}, gaming, CG, and online platform short videos datasets \textit{LIVE-YT-Gaming}~\cite{yu2022subjective}, \textit{CGVDS}~\cite{saha2023study}, \textit{KVQ}~\cite{lu2024kvq}, and \textit{FINEVD}~\cite{duan2025finevq}, as well as AIGC-oriented datasets including \textit{AGAV-3K~(test)}~(the \textbf{static quality} dimension)~\cite{cao2025agav}, \textit{AIGVQADB}~\cite{wang2024aigv}~(the \textbf{overall quality}), and \textit{GAIA}~\cite{chen2024gaia}~(the \textbf{subject quality}). For \textit{LSVQ~(train)}, we directly use the provided MOS labels as supervision. For \textit{TREND-10K}, we separately use the technical and aesthetic labels of all the $10,000$ videos and report the corresponding \textit{PLCC} and \textit{SRCC}. The results are in Tab.~\ref{tab:pretrain_inter}. Most VQA models trained on \textit{TREND-10K} achieve performance comparable to or better than those trained on \textit{LSVQ~(train)} across test datasets. The advantage is particularly evident on online short-video datasets, such as \textit{KVQ} and \textit{FINEVD}, as well as AIGC benchmarks, including \textit{AGAV3K~(test)}, \textit{AIGVQADB}, and \textit{GAIA}. These results demonstrate that \textit{TREND-10K} is an effective source training dataset for enhancing the generalization capability of commonly used VQA models.

We further study transfer learning effects on top of \textit{LSVQ(train)} pretraining. In this setting, we compare fine-tuning on \textit{TREND-10K}, \textit{KVQ (train)} ($2,926$ videos), and \textit{AIGVQADB} ($2,808$ videos), and evaluate them on conventional UGC-VQA test sets, including \textit{LSVQ (test)}~\cite{ying2021patch}, \textit{LSVQ (1080p)}, \textit{KoNViD-1K}~\cite{hosu2017konstanz}, and \textit{YT-UGC}~\cite{wang2019youtube}, as well as the content-specific OOD test sets \textit{KVQ}, \textit{FINEVD}, \textit{LIVE-YT-Gaming}, and \textit{AIGVQADB} (Tab.~\ref{tab:transfer_inter}).

\textit{TREND-10K} consistently achieves the best generalization under the \textit{LSVQ (train)}-pretrained setting. The gains are particularly notable for \textit{VITAL}, whose transfer performance surpasses all compared methods. These results indicate that fine-tuning on \textit{TREND-10K} better exploits its rich annotations and diverse content, leading to stronger and more consistent performance across diverse test scenarios.

\paragraph{Discussions on Key Experimental Points}
First, to better exploit the effects of technical and aesthetic labels, we study whether using both signals jointly leads to more effective supervision. Following the \textit{DOVER++}~\cite{wu2023exploring} setting, we directly supervise the two branches of \textit{DOVER} with technical and aesthetic scores~(given the 
inability to directly search the weighted-sum coefficients on \textit{TREND-10K}, we adopt its default coefficient settings in \textit{DOVER++}, $0.6104$ for technical and $0.3896$ for aesthetic), and compare this design with the original setting in \textit{DOVER} that uses only indirect supervision from a single target score~(results are in Tab.~\ref{tab:joint}). Joint optimization over the aesthetic and technical dimensions slightly outperforms single-dimension supervision, highlighting its effectiveness and providing a promising direction for future model development.
\begin{table}[H]
      \centering
      \tiny
      \renewcommand\arraystretch{1.1}
      \renewcommand\tabcolsep{0.3pt}
      \resizebox{\linewidth}{!}{%
      \begin{tabular}{c|c|c|c|c}
      \hline
       \multicolumn{1}{c|}{\textbf{Models}} & {\textbf{YT-UGC}} & \textbf{YT-Gaming} & \textbf{FINEVD} & \textbf{AIGVQA}\\
      \hline
      \textsc{DOVER++~(Ref.)}
        & 0.696~/~0.705 & 0.638~/~0.688 & 0.656~/~0.700 & 0.714~/~0.704 \\
      \rowcolor{light-gray0} \textsc{DOVER~(Tech.)}
        & 0.787~/~0.787 & 0.701~/~0.771 & 0.719~/~0.742 & 0.754~/~0.738\\
      \textsc{DOVER~(Aes.)}
        & 0.762~/~0.760 & 0.673~/~0.733 & 0.681~/~0.712 & 0.728~/~0.716 \\
      \rowcolor{light-gray0} \textsc{DOVER++~(All)}
        & \textbf{0.802~/~0.805}
        & \textbf{0.746~/~0.811}
        & \textbf{0.741~/~0.784}
        & \textbf{0.782~/~0.768}
       \\
       \hline
      \end{tabular}}
      \caption{Evaluation on the joint training effects.}
  \label{tab:joint}
  \end{table}

Moreover, because our data analysis shows a stable positive linear relationship between the technical and aesthetic labels within all the AIGC-trace levels, we construct training subsets by combining videos in different AIGC-trace levels~(we adopt the same configuration as in \textit{DOVER++} here, with results shown in Tab.~\ref{tab:AILEVEL}). It is evident that as the AIGC-trace level increases, the evaluation capability of trained models becomes increasingly biased toward AIGC-oriented datasets. This enables us to investigate whether perceptual-authenticity-aware subset design can improve generalization across content-specific test sets.
\begin{table}[H]
      \centering
      \tiny
      \renewcommand\arraystretch{1.2}
      \renewcommand\tabcolsep{0.15pt}
      \resizebox{\linewidth}{!}{%
      \begin{tabular}{c|c|c|c|c|c}
      \hline
       \multicolumn{1}{c|}{\textbf{Training}} &\# Videos & {\textbf{YT-UGC}} & \textbf{YT-Gaming} & \textbf{FINEVD} & \textbf{AIGVQADB}  \\
       \hline
       \textsc{L1} &$5,772$
          & 0.799~/~\textbf{0.806}
         & 0.719~/~0.753
         & 0.727~/~0.766
         & 0.741~/~0.758
         \\
     \rowcolor{light-gray0}   \textsc{L2+L3} &$2,227$
         & 0.679~/~0.665
         & 0.647~/~0.658
         & 0.573~/~0.551
         & 0.664~/~0.673
         \\
       \textsc{L4+L5} &$2,001$
         & 0.713~/~0.723
         & 0.689~/~0.734
         & 0.641~/~0.622
         & \textbf{0.788~/~0.791}
         \\
     \rowcolor{light-gray0}   \textsc{L1+L2+L3} &$7,999$
         & 0.793~/~0.801
         & 0.725~/~0.781
         & 0.721~/~0.752
         & 0.732~/~0.708
         \\
       \textsc{L1+L4+L5} &$7,773$
         & 0.776~/~0.785
         & 0.716~/~0.770
         & 0.698~/~0.766
         & 0.778~/~0.770
        \\
      \rowcolor{light-gray0}  \textsc{All} &$10,000$
         & \textbf{0.802}~/~0.805
         & \textbf{0.746~/~0.811}
         & \textbf{0.741~/~0.784}
         & 0.782~/~0.768
         \\
       \cline{1-6}
      \end{tabular}}
      \caption{Performance of training using \textit{DOVER++} setting with different AIGC-trace levels~( \textsc{L1}$\rightarrow$\textsc{L5}: low $
  \rightarrow$ high).}
   \label{tab:AILEVEL}
  \end{table}

\section{Conclusion}

We construct \textbf{TREND-10K}, a comprehensive next-generation VQA dataset grounded in user preference profiles. We also propose \textbf{TREND-Search}, which aggregates short- and long-term preference to derive principled video sampling plans, automatically crawls trend-aligned videos from \textit{YouTube}, and dynamically supplements the dataset with T2V prompts and generated AIGC videos. Our subjective study adopts three complementary dimensions to jointly characterize visual fidelity, aesthetic appeal, and perceptual AIGC-trace. Extensive experiments demonstrate that \textit{TREND-10K} serves as an effective resource for VQA pretraining and finetuning. In future, we will continue updating \textit{TREND-10K} through \textit{TREND-Search} based on new evolving trends to support \textbf{dynamic evaluation} of next-generation media perceptual quality.

{
    \bibliography{aaai2026}
}

\clearpage
\newpage

\appendix
\setcounter{secnumdepth}{2}
\section{Technical appendices and supplementary material}
\subsection{Limitations and Social Impacts}
\label{sec:Limitations}
\paragraph{Limitations}
At present, the dataset can only be distributed in the form of downloadable offline files. We have not yet developed a dedicated online distribution platform or integrated the dataset into existing online-access systems, which may introduce practical difficulties for dataset distribution and usage. The annotators of \textit{TREND-10K} are all college students aged $20$--$25$. Although this design improves annotation consistency and reduces inter-group variability, the resulting labels may not fully reflect the perceptual preferences of older user groups, potentially introducing demographic bias.

The design goal of \textit{TREND-10K} is to promote a new iteration of mainstream open-source datasets in the VQA community and thereby generate positive social impact for next-generation media quality assessment research. We plan to update the dataset periodically every three to six months in order to continuously follow evolving media trends. Each release cycle will undergo a complete subjective annotation process, allowing the corresponding dataset version to serve as an effective benchmark for perceptual video quality assessment during that period.

\paragraph{Social Impacts}
Moreover, as data from multiple release cycles are progressively accumulated, \textit{TREND-10K} is expected to gradually expand into a large-scale benchmark, with further improvements in training effectiveness and cross-domain generalization capability. Owing to the highly automated design of \textit{TREND-Search}, the entire data collection pipeline requires only minimal human intervention, making scheduled large-scale data acquisition practical and efficient. In future work, we plan to open-source the \textit{TREND-Search} framework, the annotation platform, and each dataset release to further contribute to the open-source VQA community.

\subsection{Summary of Existing Datasets}
\label{sec:Summary}
We summarize representative datasets for UGC/PGC-VQA and AIGC-VQA in Tab.~\ref{tab:VQA_summary} and~\ref{tab:AIGC_VQA_summary}, respectively.
\begin{table*}[t]
    \centering
    \renewcommand\arraystretch{1.2}
    \renewcommand\tabcolsep{3pt}
    
    \vspace{-6pt}
    \belowrulesep=0pt\aboverulesep=0pt
    \resizebox{\textwidth}{!}{%
    \rowcolors{2}{gray!12}{white}
    \begin{tabular}{l|c|c|c|c}
        \hline
        \textit{Datasets for UGC/PGC-VQA} & Year & \# Videos & \# MOS & Description \\ 
        \hline
        LIVE-VQA & 2010 & 160 & 160 & Full-reference video quality rating  \\ 
        CVD2014 & 2016 & 234 & 234 & Quality assessment of video captured by cameras \\ 
        LIVE-Qualcomm & 2017 & 208 & 208 & Mobile in-capture video quality rating \\ 
        KoNViD-1K & 2017 & 1,200 & 1,200 & Unified UGC video quality rating \\ 
        LIVE-NFLX-I & 2017 & 558 & 558 & Quality-of-experience~(QoE) rating of hand-craft streaming videos \\ 
        LIVE-VQC & 2018 & 585 & 585 & Quality rating of real world UGC videos \\ 
        LIVE-NFLX-II & 2018 & 420 & 420 & QoE rating of real-world streaming videos\\ 
        WaterlooSQoE-III & 2018 & 450 & 450 & QoE rating of hand-craft streaming videos\\ 
        LBVD & 2019 & 1,013 & 1,013 & QoE assessment of in-the-wild streaming videos \\ 
        YouTube-UGC & 2020 & 1,380 & 1,380 & Quality rating of UGC videos \\ 
        WaterlooSQoE-IV & 2020 & 1,350 & 1,350 & Large-scale QoE assessment of hand-craft streaming videos \\ 
        CGVDS & 2020 & 360 & 360 & Quality assessment of cloud gaming videos with encoding distortions \\ 
        LSVQ & 2021 & 39,075 & 39,075 & Large-scale quality rating of UGC videos \\ 
        LIVE-YT-Gaming & 2022 & 600 & 600 & Quality assessment of real UGC gaming videos from YouTube \\ 
        TaoLive & 2023 & 3,762 & 3,762 & Quality rating of live streaming~(compressed) videos \\ 
        Maxwell & 2023 & 4,543 & 9,086 & Fine-grained~(technical/aesthetic) quality rating of UGC videos \\ 
        SJTU-UAV & 2023 & 520 & 520 & Audio-visual quality assessment of in-the-wild UGC A/V sequences \\
        KVQ & 2024 & 4,200 & 4,200 & Quality assessment of short-form UGC videos  \\
        SR4KVQA & 2024 & 600 & 600 & Quality assessment of 4K super-resolution distorted videos \\
        FineVD & 2024 & 6,104 & 36,624 & Fine-grained UGC video quality assessment \\
        OAVQAD+ & 2024 & 625 & 625 & Audio-visual quality assessment of omnidirectional videos \\
        OmniVQA-MOS-20K & 2025 & 20,000 & 20,000 & Large-scale MIDB for quality rating for in-the-wild UGC videos \\ 
        \hline
    \end{tabular}%
    }
    \caption{Summary of existing UGC/PGC VQA MIDBs.}
    \label{tab:VQA_summary} 
\end{table*}
    \vspace{-6pt}

\begin{table*}[t]
    \centering
    \renewcommand\arraystretch{1.2}
    \renewcommand\tabcolsep{3pt}
    \belowrulesep=0pt\aboverulesep=0pt
    
    \vspace{-6pt}
    \resizebox{\textwidth}{!}{%
    \rowcolors{2}{gray!12}{white}
    \begin{tabular}{l|c|c|c|c}
        \hline
        \textit{Datasets for AIGC VQA} & Year & \# Videos & \# MOS & Description \\ 
        \hline
        T2VQA-DB & 2024 & 10,000 & 10,000 & Subjective-aligned quality assessment of text-to-video generated videos \\ 
        LGVQ & 2024 & 1,800 & 5,400 & Multi-dimensional quality assessment of AIGC videos \\ 
        AIGVQA-DB & 2024 & 2,808 & 8,424  & Large-scale perceptual quality assessment of AI-generated videos  \\ 
        AGAVQA-3k & 2024 & 3,382 & 10,146 & Multi-dimensional audio-visual quality assessment \\ 
        GAIA & 2024 & 9,180 & 9,180 & Action quality assessment of AI-generated videos \\ 
        TDVE-DB & 2025 & 3,857 & 11,571 & Quality assessment of text-driven video editing  \\ 
        AIGVE-60K & 2025 & 58,500 & 120,000 & Large-scale benchmark for AI-generated video  \\ 
        \hline
    \end{tabular}%
    }
    \caption{Summary of existing AIGC VQA datasets.}
    \vspace{-6pt}
    \label{tab:AIGC_VQA_summary} 
\end{table*}
\label{AIGC_VQA_summary}

\subsection{Composition statistics of the TREND-10K dataset}
\label{sec:dataset_composition}
Tab.~\ref{tab:dataset_composition} summarizes the composition of TREND-10K, including the video sources and sample counts for each dataset component.
\begin{table}[H]
  \centering
  \footnotesize 
  \setlength{\tabcolsep}{6pt}
  \renewcommand{\arraystretch}{1.1}
  \begin{tabular}{@{}llr@{}}
    \toprule
    \textbf{Dataset part} & \textbf{Video source} & \textbf{\# Videos} \\
    \midrule
    \multirow{7}{*}{Trend-driven part (7000)}
      & YouTube 2025.10 & 588 \\
      & YouTube 2025.11 & 667 \\
      & YouTube 2025.12 & 675 \\
      & YouTube 2026.1  & 1,023 \\
      & YouTube 2026.2  & 1,125 \\
      & YouTube 2026.3  & 1,860 \\
      & Supplementary T2V & 1,062 \\
    \midrule
    \multirow{2}{*}{Static part (3000)}
      & \textit{VITAL-Pretrain} & 2,200 \\
      & \textit{AIGVE-60K} & 800 \\
    \midrule
    \textbf{Total} & -- & \textbf{10,000} \\
    \bottomrule
  \end{tabular}
  \caption{Composition of the TREND-10K dataset.}
  \label{tab:dataset_composition}
\end{table}

\subsection{LLM Usage Justification}
\label{sec:LLM_Justification}
The LLM is used only for writing, editing, or formatting purposes and does \emph{not} impact the core methodology, scientific rigor, or originality of the research.

\subsection{Details for the Subjective Experiment Online Platform}
\label{sec:experiment_online_platform}
We present a large-scale crowdsourced annotation platform designed to support subjective qualityassessment of AI-generated video content. The platform facilitates the collection of human perceptual judgments across three orthogonal   
  dimensions: technical quality, aesthetic quality, and AI trace intensity — the last capturing the degree to which a video is perceived as synthetically generated.                                                                            
                                            
  To ensure annotation quality, all annotators must complete a structured training phase consisting of 32 representative videos with calibration guidance before accessing the formal annotation task. Each annotation session presents      
  videos with synchronized playback controls and prompts the annotator to provide integer ratings on a 1–5 Likert scale for each dimension, along with an optional free-text remark. Submission is atomic: all three dimension scores are
  recorded together with a UTC timestamp, enabling temporal analysis of annotation behavior.                                                                                                                                                 
                  
\begin{figure*}[h!]
        \centering
        \includegraphics[width=0.95\linewidth]{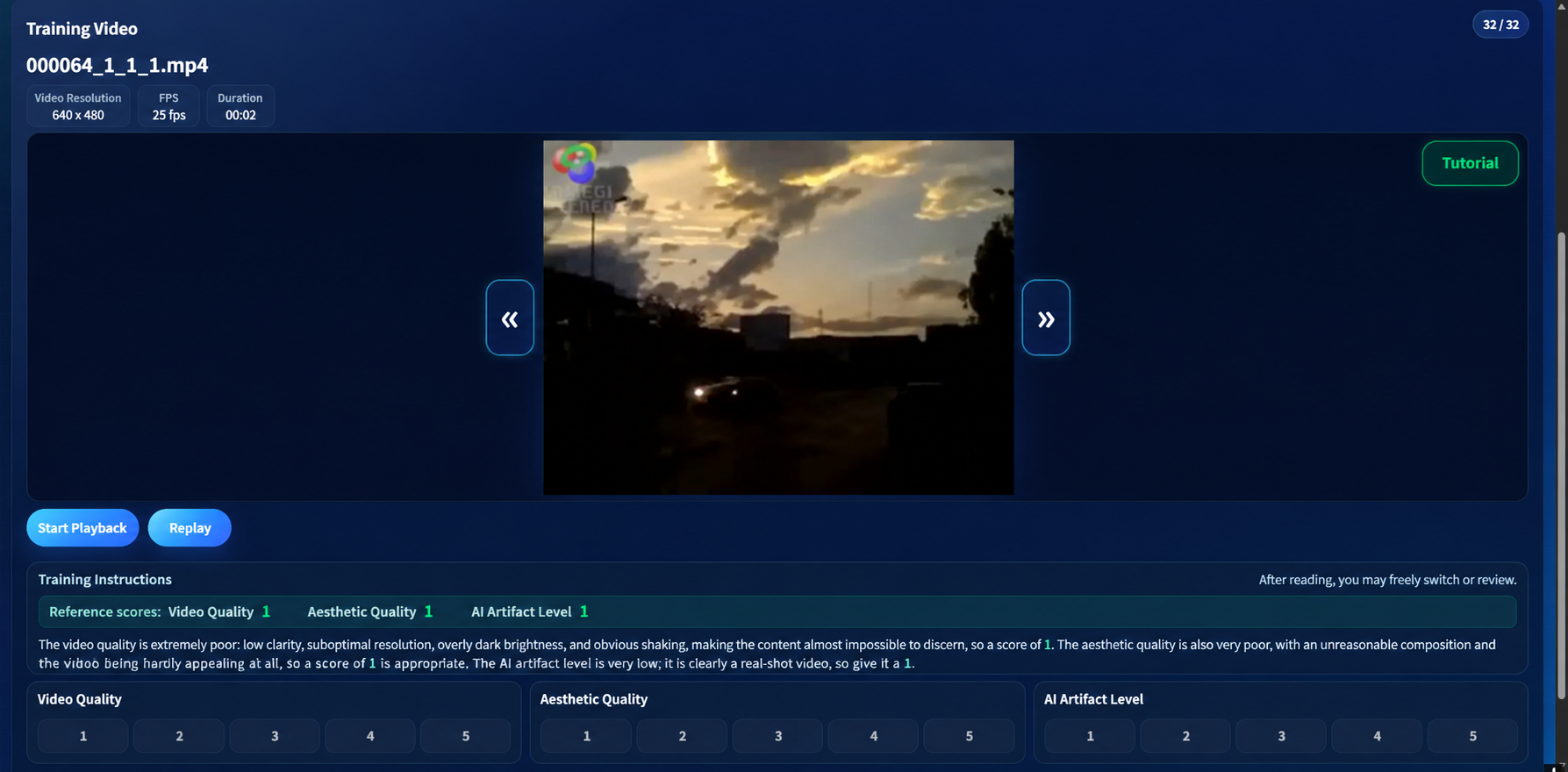}
        \caption{Training phase}
        \label{fig:traning_phase}
  \end{figure*}

\begin{figure*}[h!]
        \centering
        \includegraphics[width=0.95\linewidth]{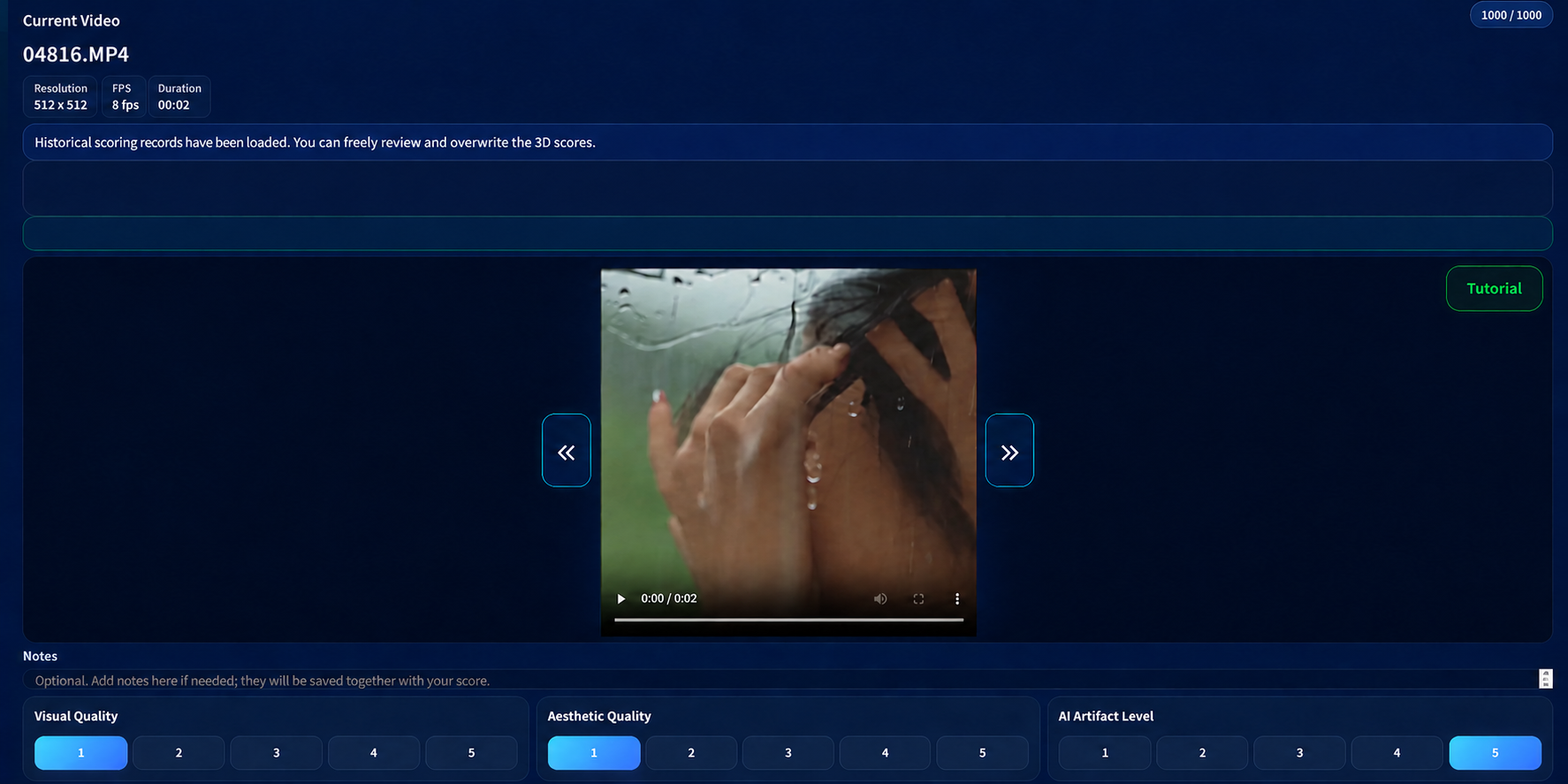}
        \caption{Annotating phase}
        \label{fig:annotating_phase}
  \end{figure*}

  The backend is implemented as a FastAPI application with a PostgreSQL persistence layer, served via Nginx with X-Accel-Redirect for efficient authenticated video streaming. Session management is cookie-based with configurable TTL, and 
  all video file access is mediated through internal Nginx location blocks, preventing direct URL-based enumeration.
  Fig.~\ref{fig:traning_phase} and~\ref{fig:annotating_phase} illustrate the user interface of the proposed annotation platform.

\subsection{Subjective Experiment Guideline}
\label{sec:guideline}
\begin{figure*}[p]
        \centering
        \includegraphics[width=0.8\linewidth, page=1]{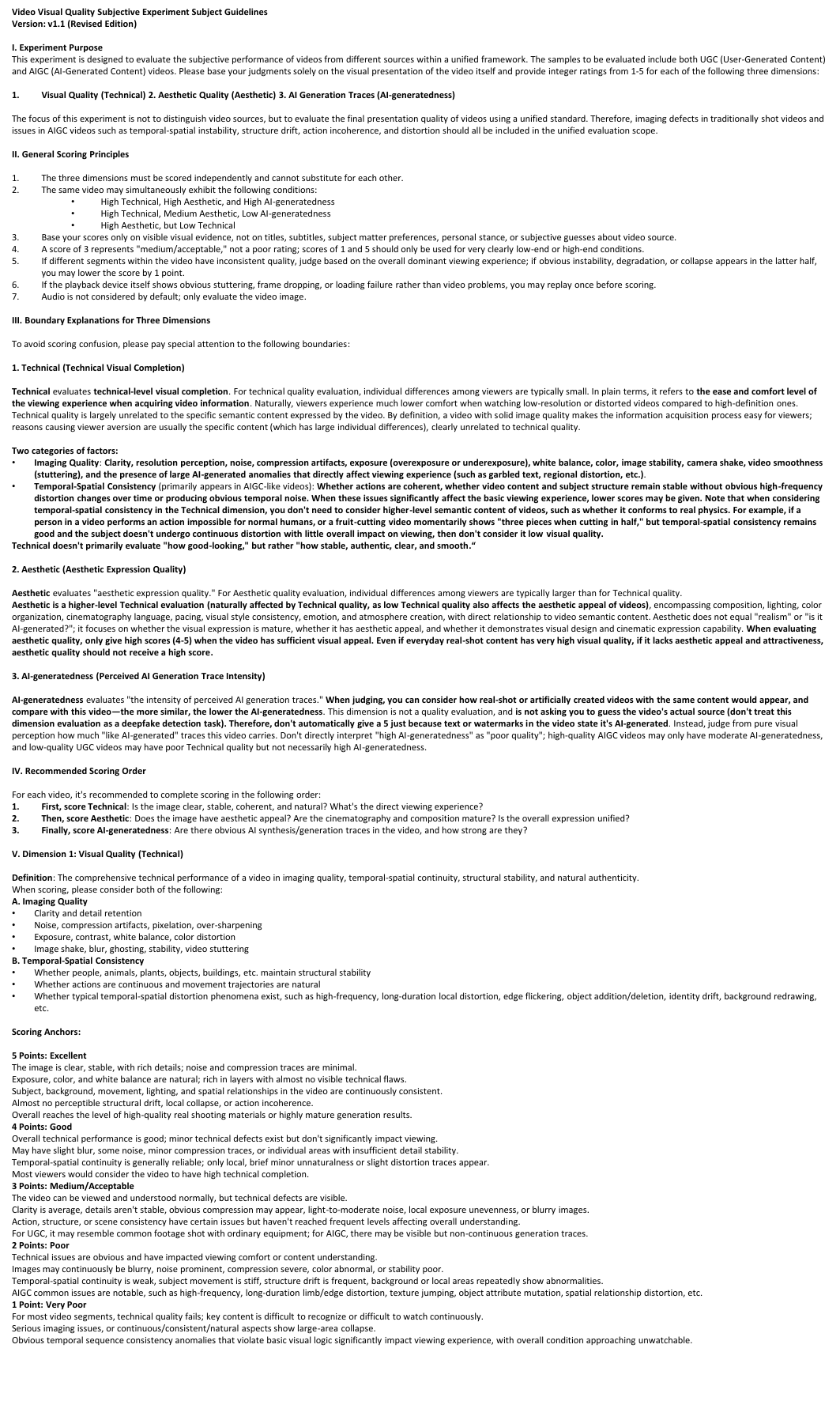}
         \caption{Subjective Experiment Guideline, part 1}
         \label{fig:guideline1}
    \end{figure*}

\begin{figure*}[p]
        \centering
        \includegraphics[width=0.8\linewidth, page=2]{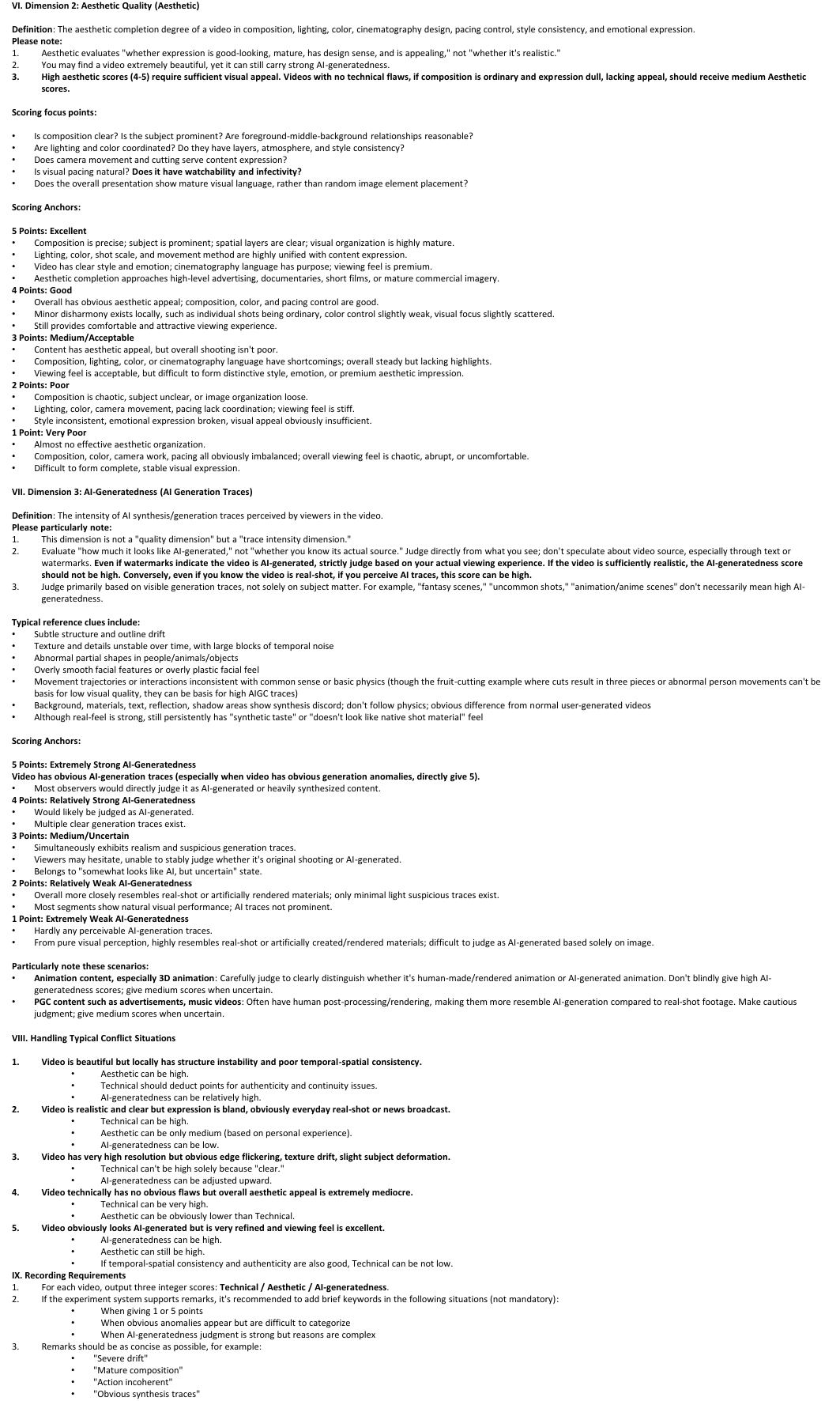}
         \caption{Subjective Experiment Guideline, part 2}
         \label{fig:guideline2}
    \end{figure*}

Fig.~\ref{fig:guideline1} and~\ref{fig:guideline2} present the complete guideline used in our subjective annotation experiments.
\makeatletter
\setlength{\@fptop}{-10pt}
\makeatother

\subsection{Details for Trending Profile and Sampling Plan Construction Process}
\label{sec:D1}
\begin{figure*}[p]
    \centering
    \includegraphics[width=0.75\linewidth]{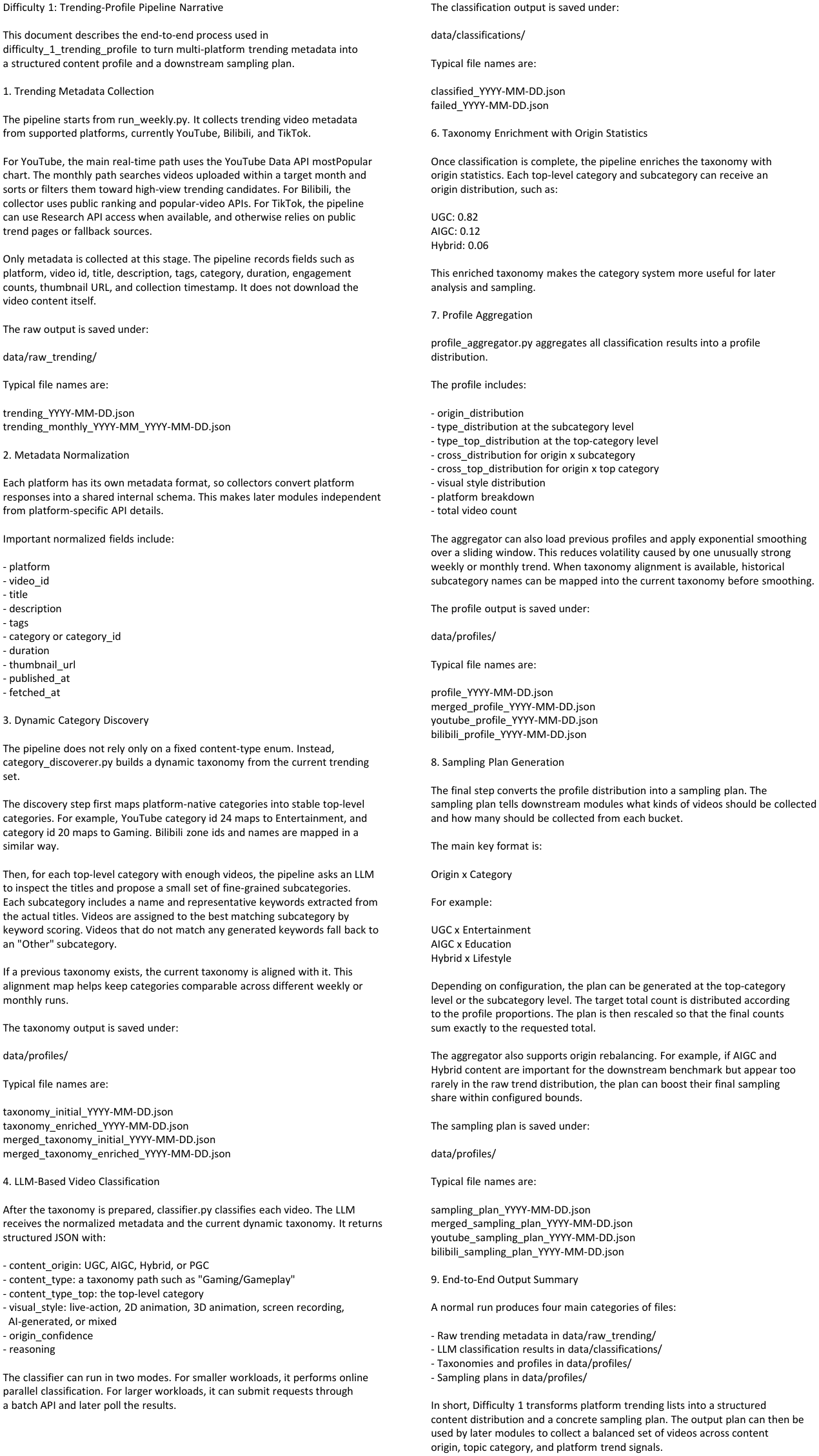}
    \caption{Details for trending profile and sampling plan construction process, part 1}
    \label{fig:trend_profile_1}
\end{figure*}

\begin{figure*}[p]
    \centering
    \includegraphics[width=0.7\linewidth]{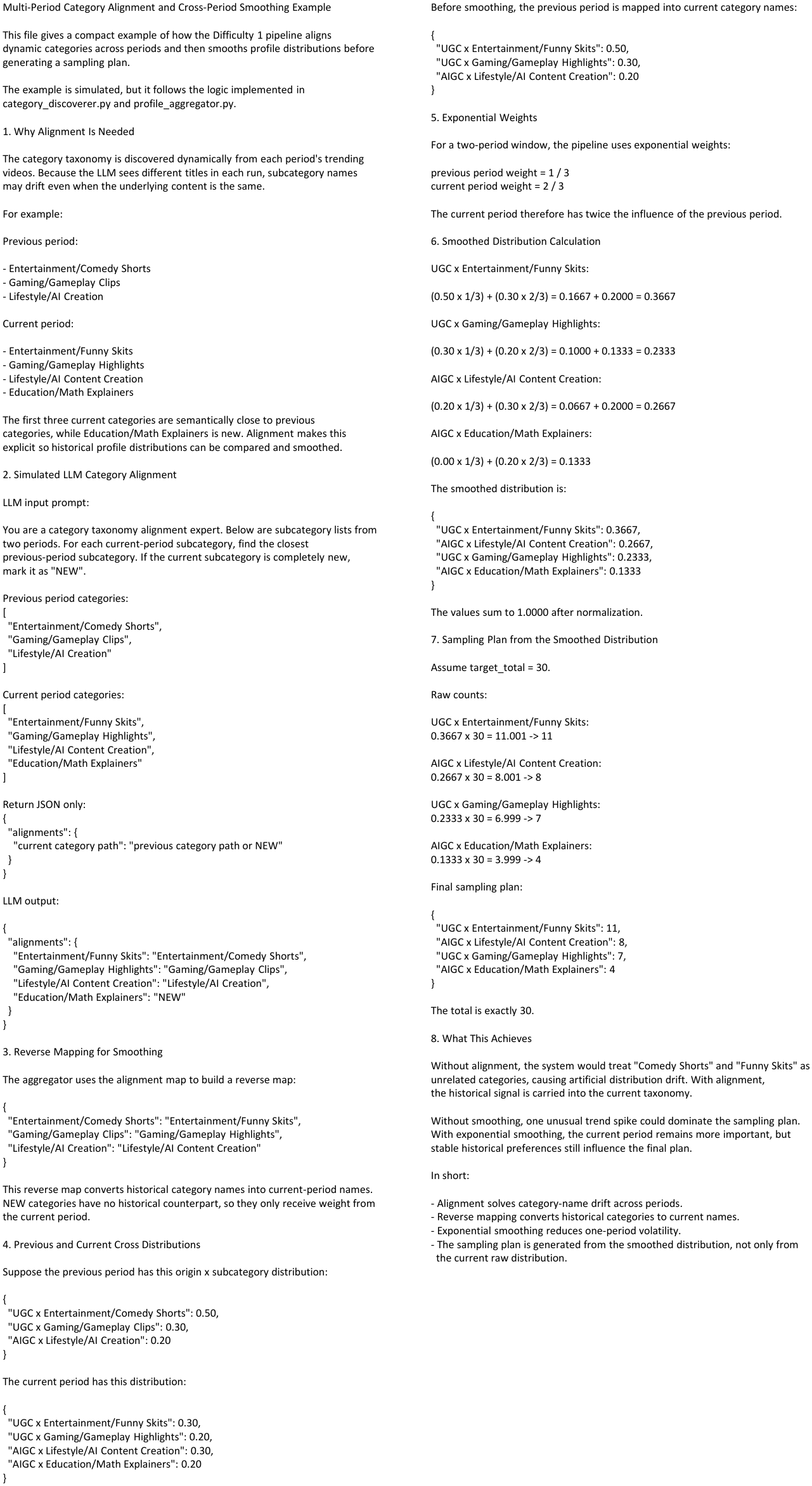}
    \caption{Details for trending profile and sampling plan construction process, part 2}
    \label{fig:trend_profile_2}
\end{figure*}

Fig.~\ref{fig:trend_profile_1} and~\ref{fig:trend_profile_2} illustrate the construction process of the trending profile and the corresponding sampling plan used in TREND-Search.
\makeatletter
\setlength{\@fptop}{-10pt}
\makeatother

\subsection{Details for Supplementary T2V Generation Models and Prompts}
\label{sec:T2Vmodels}
\begin{figure*}[b]
    \centering
    \includegraphics[width=0.9\linewidth]{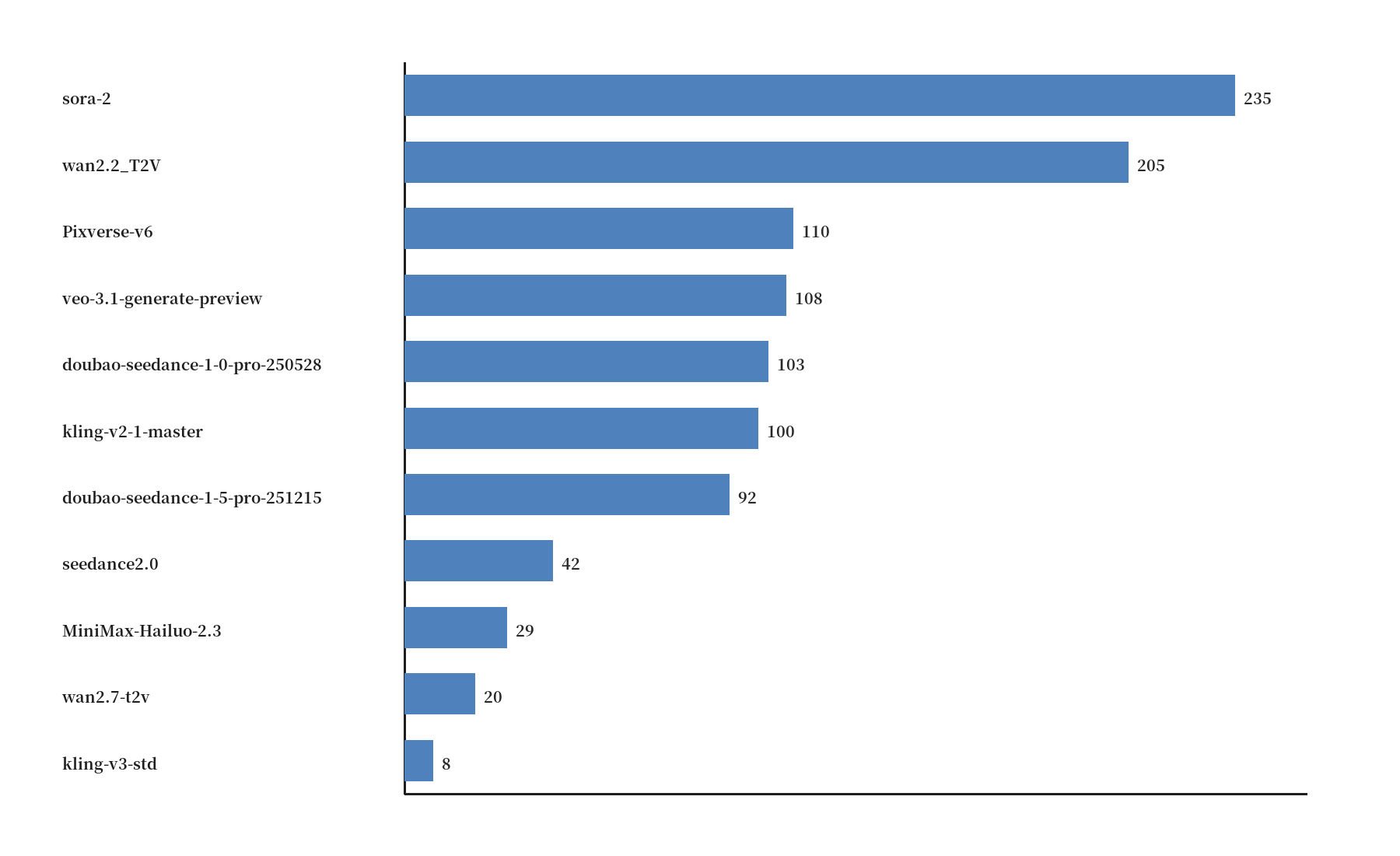}
    \caption{Models included in this study}
    \label{fig:t2v_models}
\end{figure*}

\begin{figure*}[p]
    \includegraphics[width=0.8\linewidth]{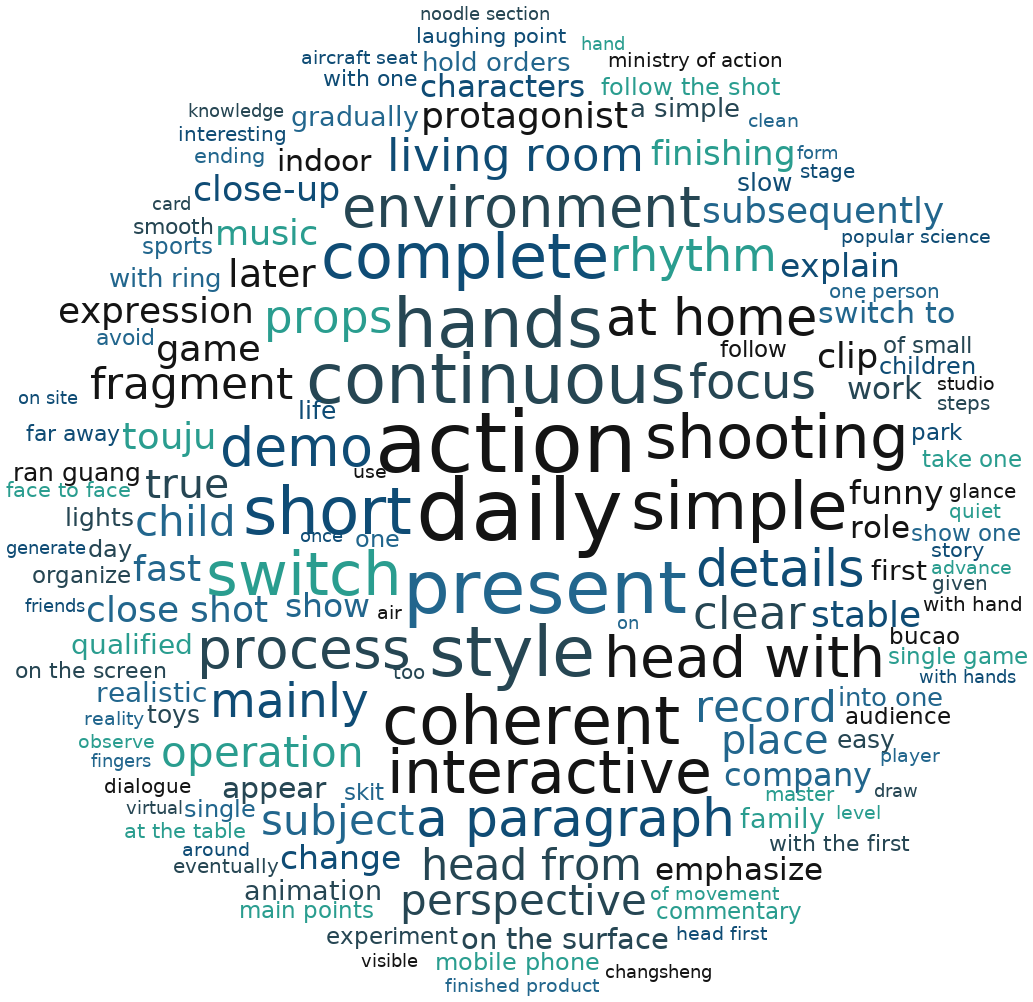}
    \caption{Word cloud for supplementary T2V generation prompts.}
    \label{fig:t2v_wordcloud}
\end{figure*}

\begin{figure*}[p]
    \centering
    \includegraphics[width=\linewidth]{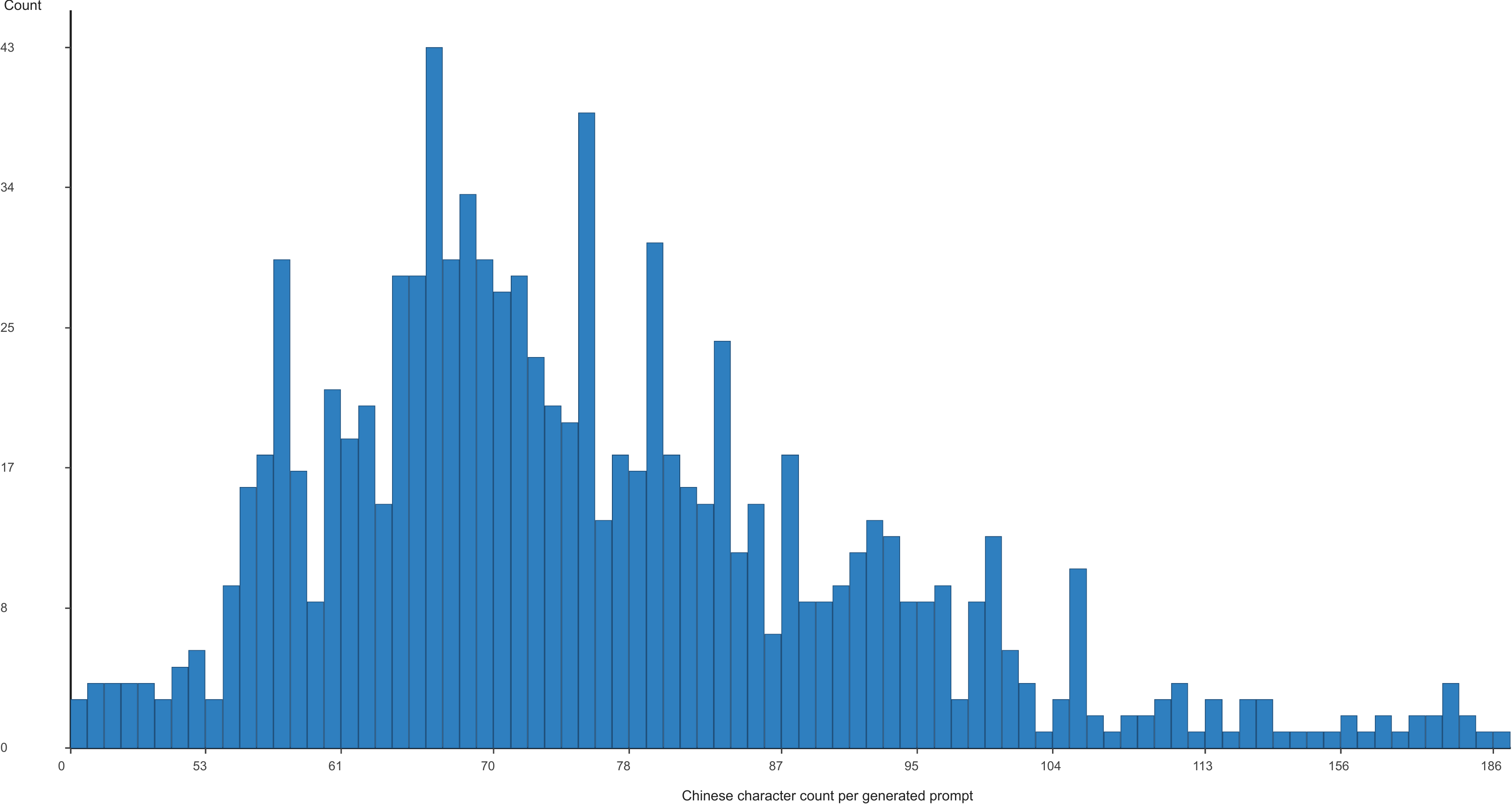}
    \caption{Histogram of Word Count in generated prompts}
    \label{fig:prompt_length}
\end{figure*}

\begin{figure*}[p]
    \centering
    \includegraphics[width=\linewidth]{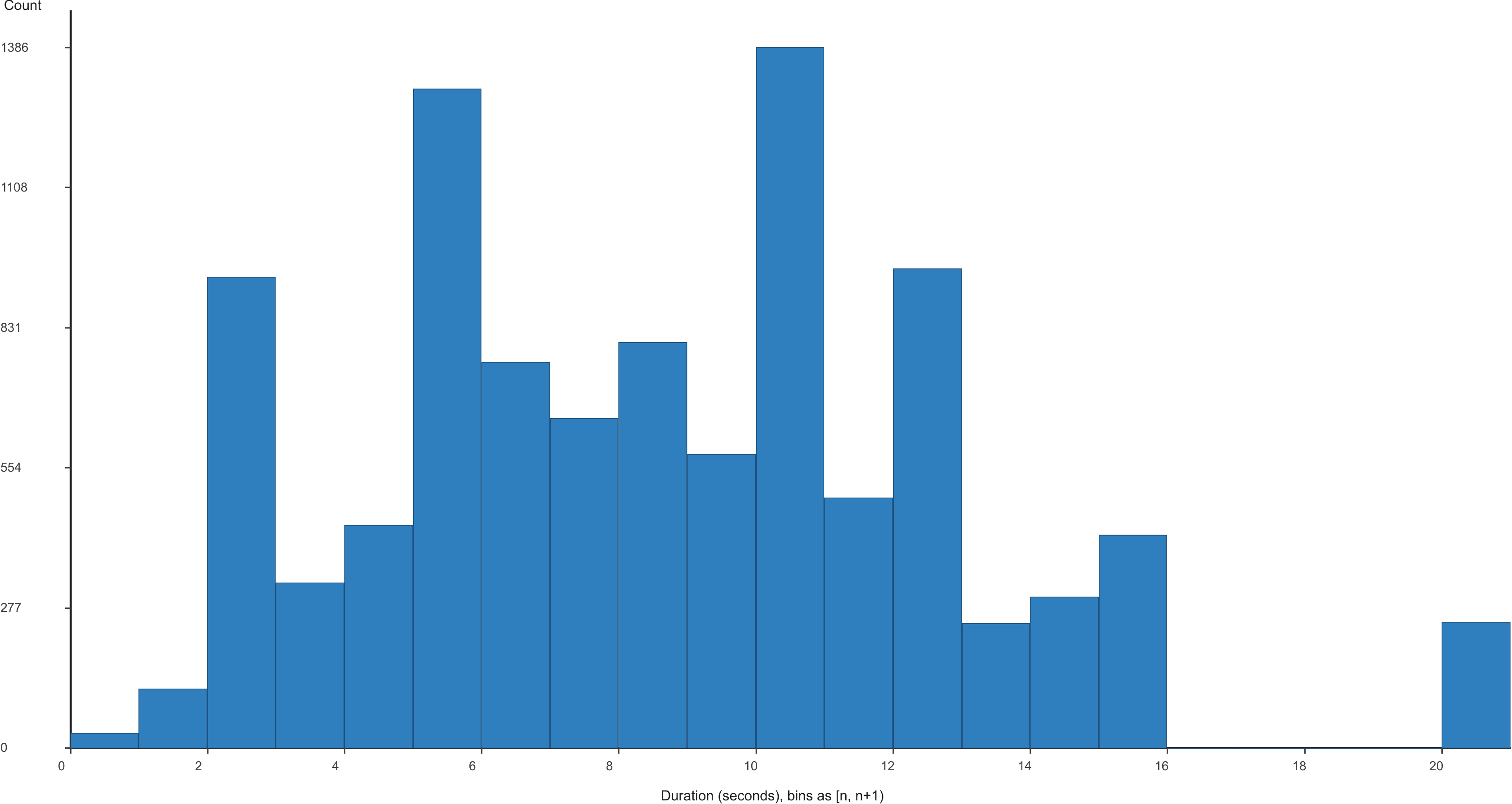}
    \caption{All videos duration histogram}
    \label{fig:video_duration}
\end{figure*}

\begin{figure*}[p]
    \centering
    \includegraphics[width=\linewidth]{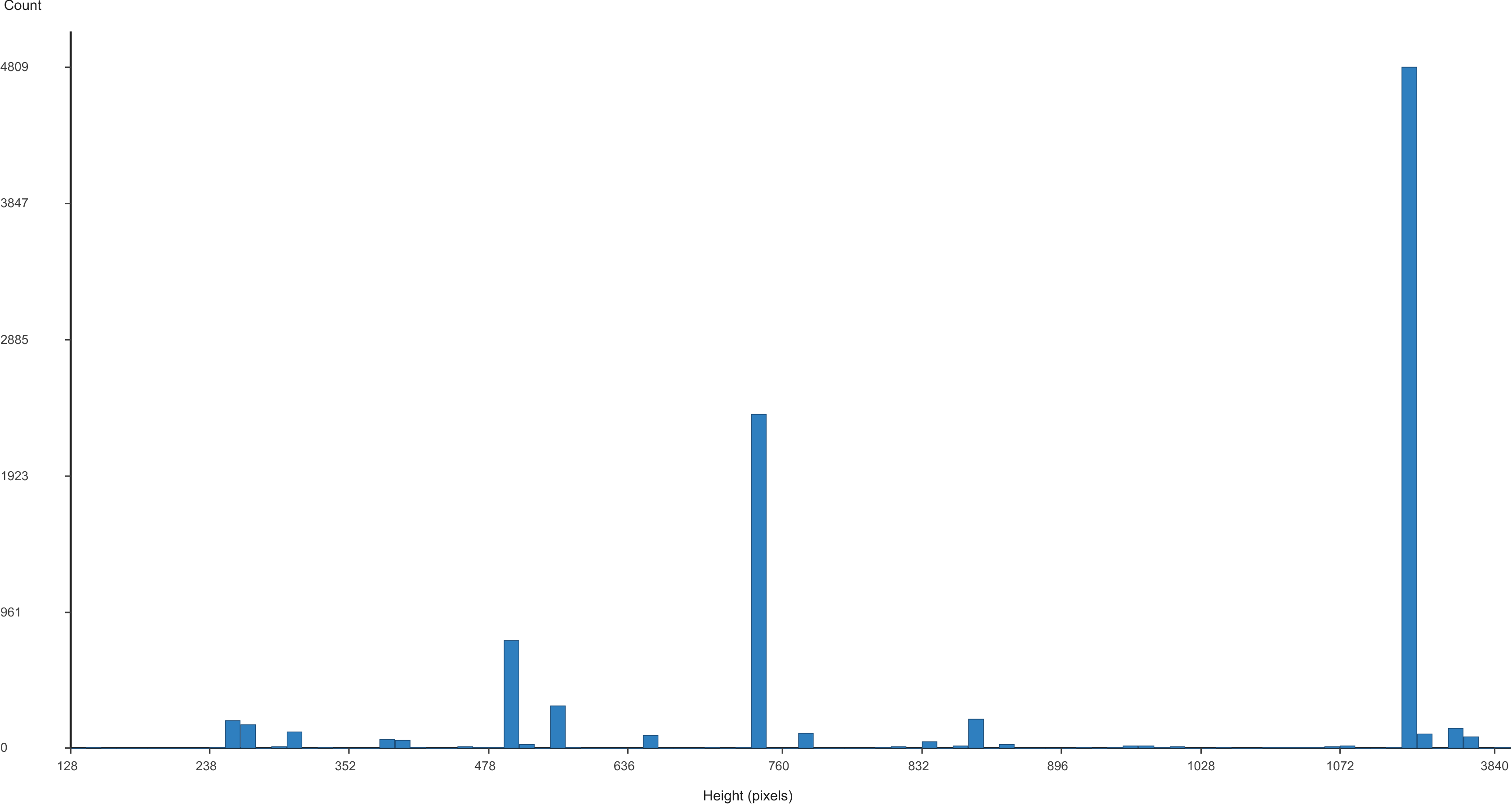}
    \caption{All videos height histogram}
    \label{fig:height_histogram}
\end{figure*}

\begin{figure*}[p]
    \centering
    \includegraphics[width=\linewidth]{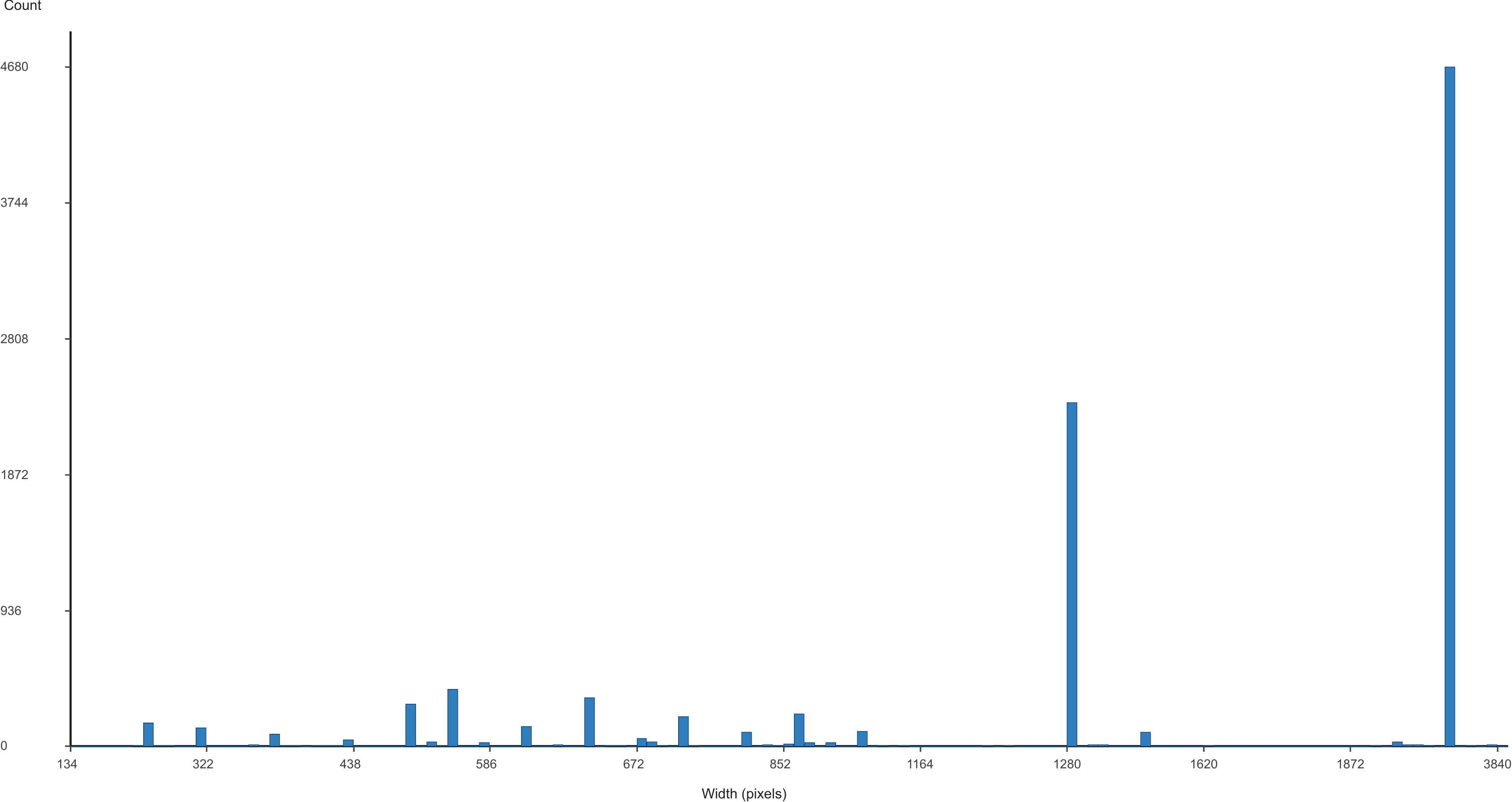}
    \caption{All videos width histogram}
    \label{fig:width_histogram}
\end{figure*}

\begin{figure*}[p]
    \centering
    \includegraphics[width=\linewidth]{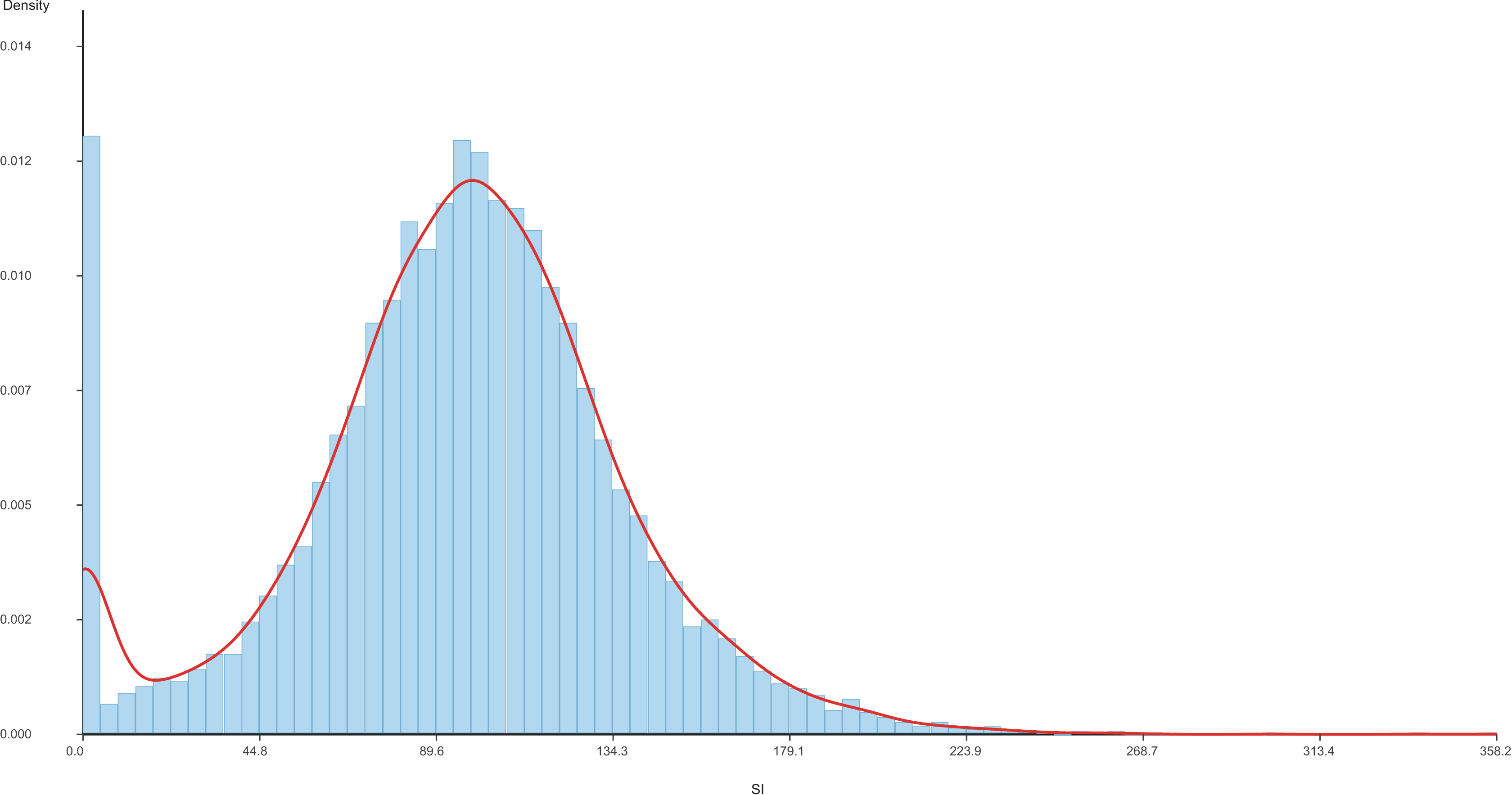}
    \caption{Fitted curve for SI}
    \label{fig:SI}
\end{figure*}

\begin{figure*}[p]
    \centering
    \includegraphics[width=\linewidth]{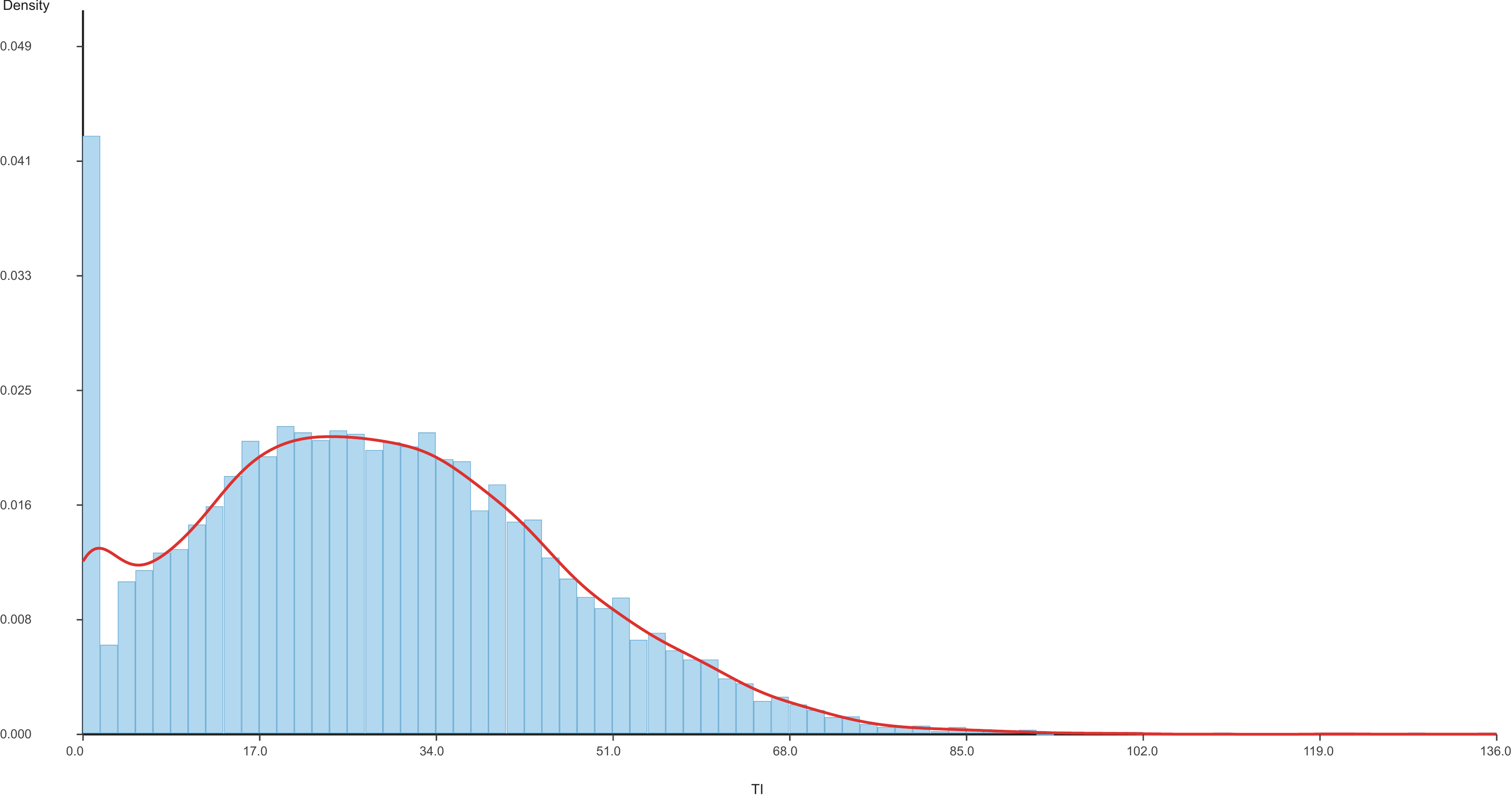}
    \caption{Fitted curve for TI}
    \label{fig:TI}
\end{figure*}

  \begin{itemize}
  \item \textbf{Sora-2}: Sora-2 is an OpenAI video+audio generation model, officially announced on 2025-09-30; it is closed-source, and its official page is \url{https://openai.com/index/sora-2/}.

  \item \textbf{Wan2.2\_T2V}: Wan2.2\_T2V is Alibaba's Wan 2.2 text-to-video model; a public repository was available by 2026-03~(with no clearly announced exact launch date), it is open-source under Apache-2.0, and the
  official repository is \url{https://github.com/Wan-Video/Wan2.2}.

  \item \textbf{PixVerse-v6}: PixVerse-v6 is the PixVerse V6 video generation model, released on 2026-03-30 according to the official blog; it is closed-source, and the official page is \url{https://pixverse.ai/en/blog/pixverse-launches-v6-advancing-ai-video-generation}.

  \item \textbf{Veo-3.1-Generate-Preview}: Veo-3.1-Generate-Preview is Google's Veo 3.1 preview video model, with a documented release date of 2025-10-15; it is closed-source, and the official documentation page is
  \url{https://cloud.google.com/vertex-ai/generative-ai/docs/models/veo/3-1-generate-preview}.

  \item \textbf{Doubao-Seedance-1.0-Pro-250528}: Doubao-Seedance-1.0-Pro-250528 is a ByteDance Seedance 1.0 internal/API variant; the Seedance 1.0 paper was published on 2025-06-11, while this exact variant has no
  separate public launch date, it is closed-source, and the official reference is \url{https://seed.bytedance.com/zh/public_papers/seedance-1-0-exploring-the-boundaries-of-video-generation-models}.

  \item \textbf{Kling-v2-1-master}: Kling-v2-1-master is a Kling 2.1 series API/internal model ID; no separate public launch date is available for this exact ID, while the first public Kling model launch was on 2024-06-
  10, it is closed-source, and the official reference is \url{https://ir.kuaishou.com/news-releases/news-release-details/kuaishou-unveils-proprietary-video-generation-model-kling}.

  \item \textbf{Doubao-Seedance-1.5-Pro-251215}: Doubao-Seedance-1.5-Pro-251215 is a ByteDance Seedance 1.5 Pro audio-video generation model, officially released on 2025-12-16; it is closed-source, and the official page
  is \url{https://research.doubao.com/en/blog/sound-and-vision-all-in-one-take-the-official-release-of-seedance-1-5-pro}.

  \item \textbf{Seedance2.0}: Seedance2.0 is ByteDance's Seedance 2.0 video generation model, officially launched on 2026-02-12; it is closed-source, and the official page is \url{https://seed.bytedance.com/en/blog/seedance-2-0-official-launch}.

  \item \textbf{MiniMax-Hailuo-2.3}: MiniMax-Hailuo-2.3 is the MiniMax Hailuo 2.3 model, announced on 2025-10-28; it is closed-source, and the official page is \url{https://www.minimax.io/news/minimax-hailuo-23}.

  \item \textbf{Wan2.7-T2V}: Wan2.7-T2V is Alibaba Cloud's Wan2.7 video generation model, launched on 2026-04-07; it is a closed-source service model, and the official page is \url{https://www.alibabacloud.com/blog/603009}.

  \item \textbf{Kling-v3-std}: Kling-v3-std is a Kling 3.0 standard variant API model ID, corresponding to the official Kling 3.0 launch on 2026-02-05; it is closed-source, and the official reference is \url{https://ir.kuaishou.com/news-releases/news-release-details/kling-ai-launches-30-model-ushering-era-where-everyone-can-be}.
  \end{itemize}
  Fig.~\ref{fig:t2v_models} and~\ref{fig:t2v_wordcloud} summarize the T2V models used for supplementary video generation and the lexical distribution of the corresponding prompts.
  Fig.~\ref{fig:prompt_length} and~\ref{fig:video_duration} present the distributions of prompt length and generated video duration for the supplementary T2V data, respectively.
  Fig.~\ref{fig:height_histogram} and~\ref{fig:width_histogram} demonstrate the distribution of lengths and heights of all videos used for annotation in the study. 
  Fig.~\ref{fig:SI} and~\ref{fig:TI} show the fitted curve of SI and TI of all videos.

\subsection{Details for Online Search and Video Crawling Process}
\label{sec:D2}
\begin{figure*}[p]
        \centering
        \includegraphics[width=0.6\linewidth]{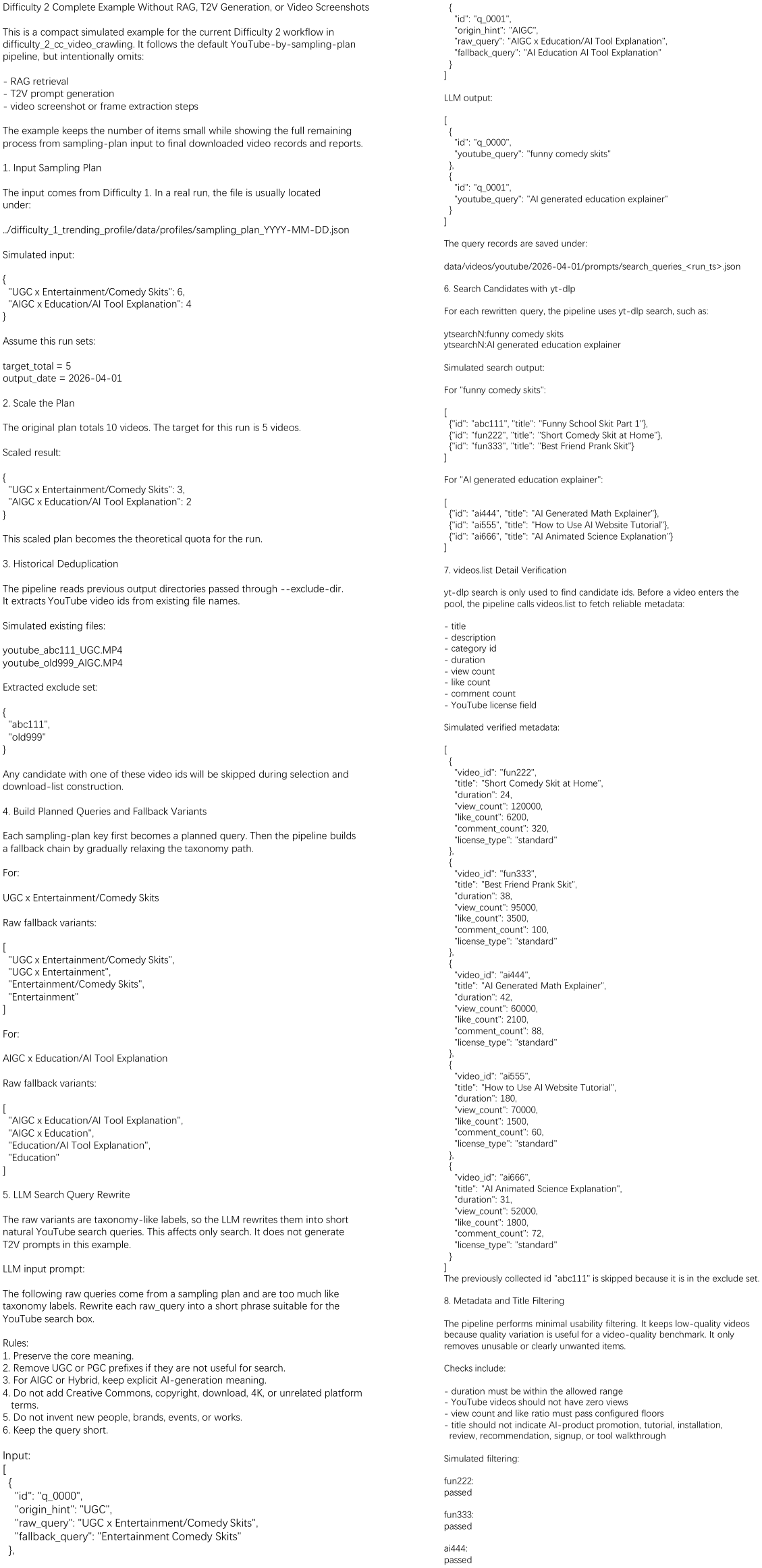}
        \caption{Details for Online Search and Video Crawling Process, part 1.}
        \label{fig:online_search_1}
    \end{figure*}

\begin{figure*}[p]
        \centering
        \includegraphics[width=0.7\linewidth]{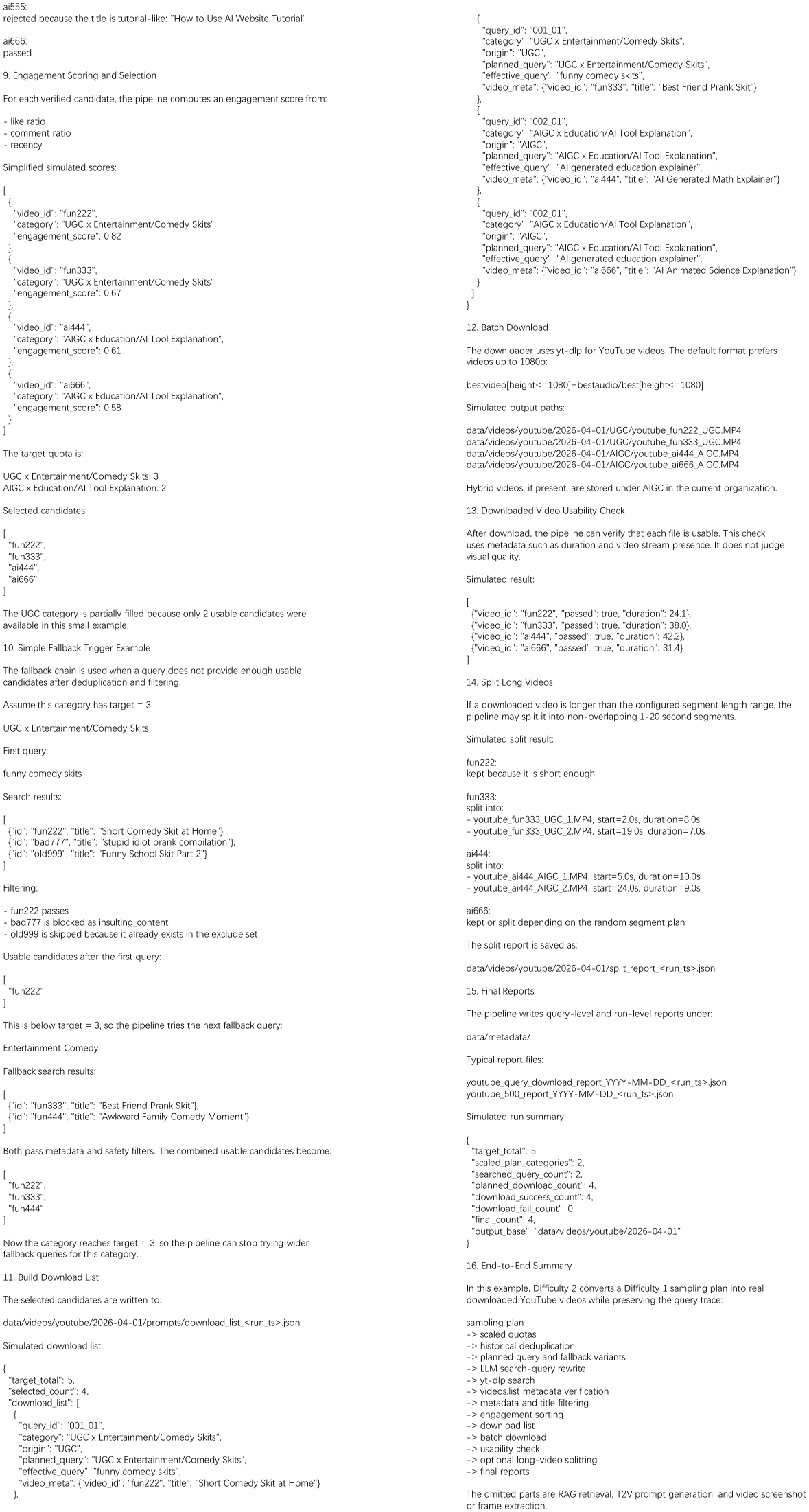}
        \caption{Details for Online Search and Video Crawling Process, part 2.}
        \label{fig:online_search_2}
    \end{figure*}

Fig.~\ref{fig:online_search_1} and~\ref{fig:online_search_2} illustrate the complete workflow of the online search and video crawling process, including query generation, candidate retrieval, metadata verification, and video collection.

\subsection{Details for T2V Generation Process}
\label{sec:D3}
\begin{figure*}[p]
        \centering
        \includegraphics[width=0.72\linewidth]{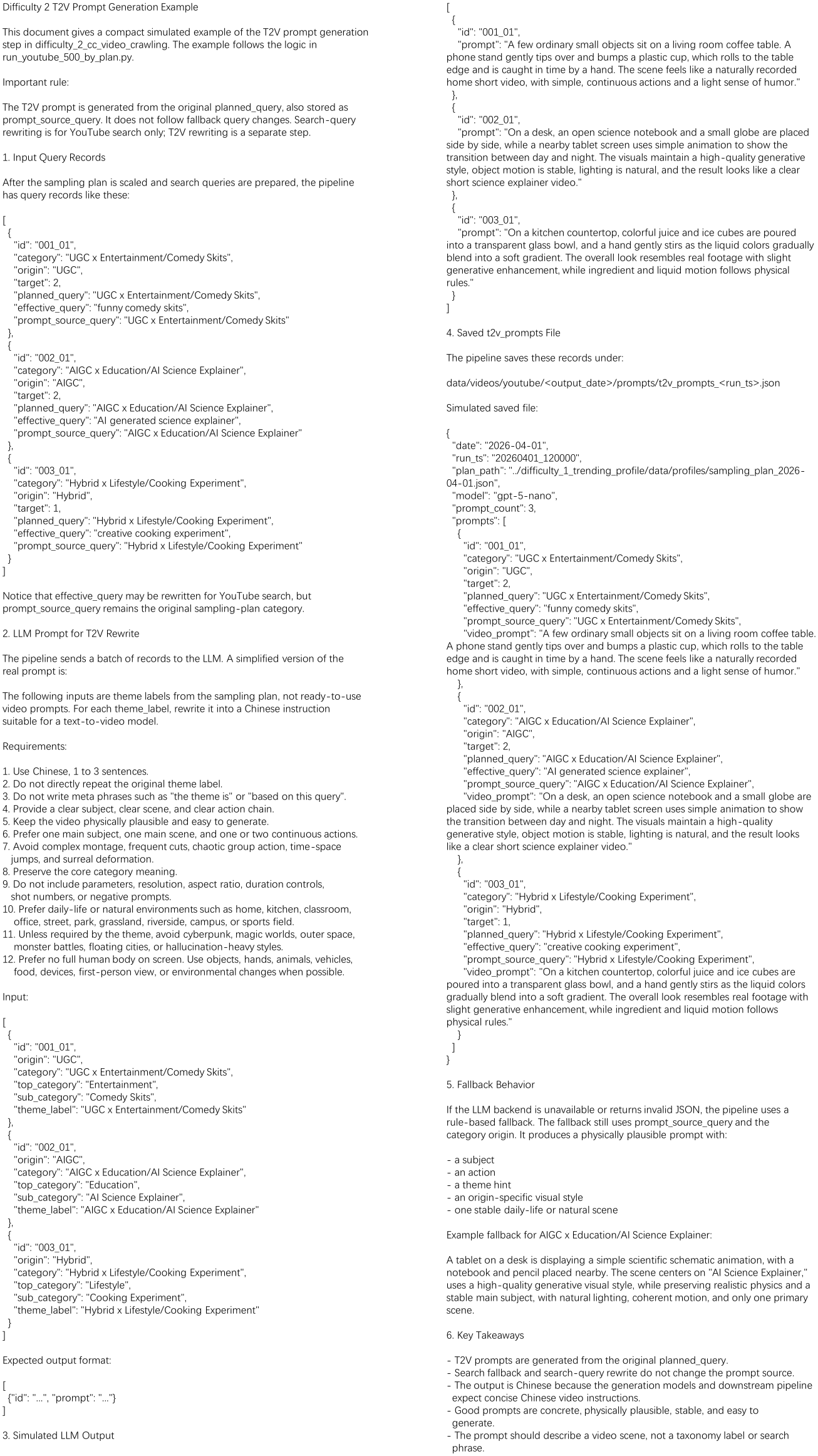}
        \caption{Details for T2V Generation Process.}
        \label{fig:t2v_generation}
    \end{figure*}

Fig.~\ref{fig:t2v_generation} provides a detailed illustration of the T2V generation process, including prompt construction, prompt refinement, video generation, and quality verification.

\subsection{Visualization and More Statistical Analysis for TREND-10K Dataset}
\label{sec:Visualization_and_More_Statistical_Analysis}
\begin{figure*}[p]
    \centering
    \includegraphics[width=0.9\linewidth]{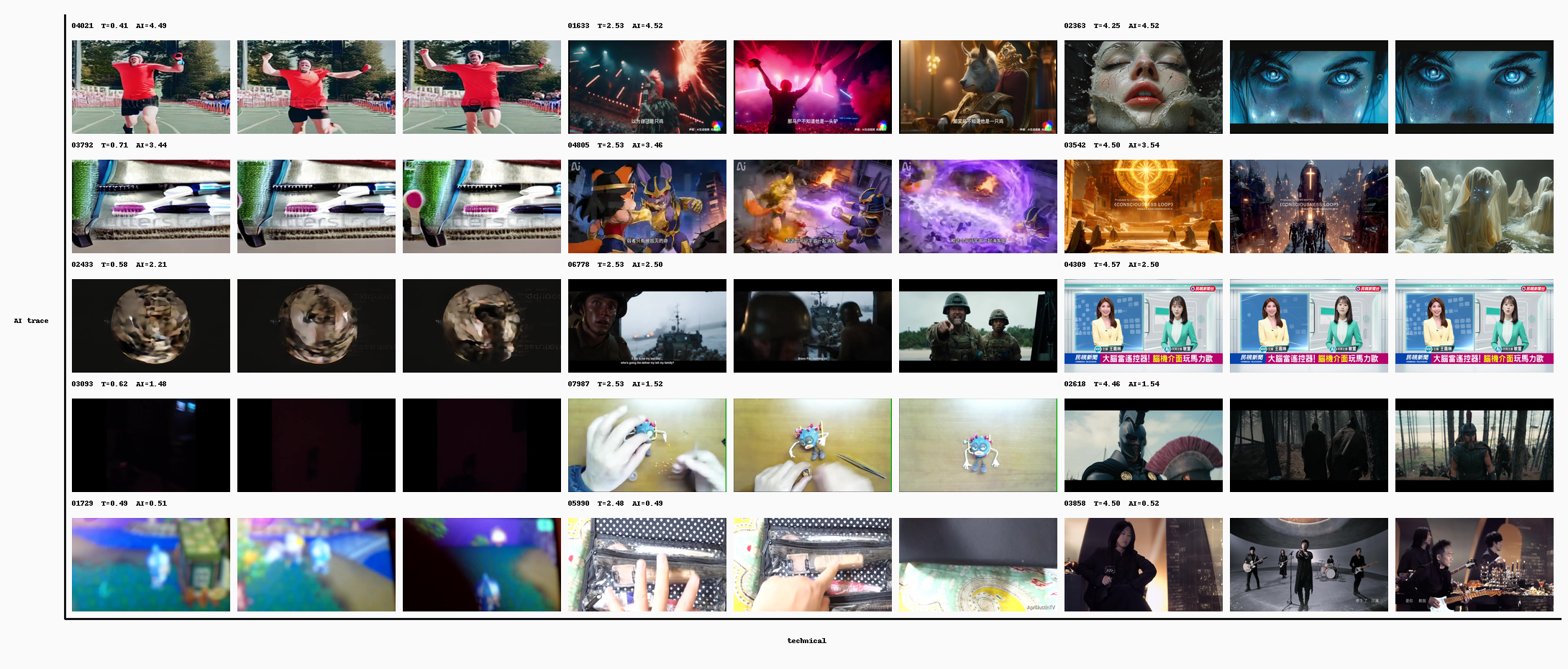}
    \caption{Visualization for examples of multiple AI trace-Technical dimensions levels}
    \label{fig:aitrace-technical}
\end{figure*}

\begin{figure*}[p]
    \centering
    \includegraphics[width=0.9\linewidth]{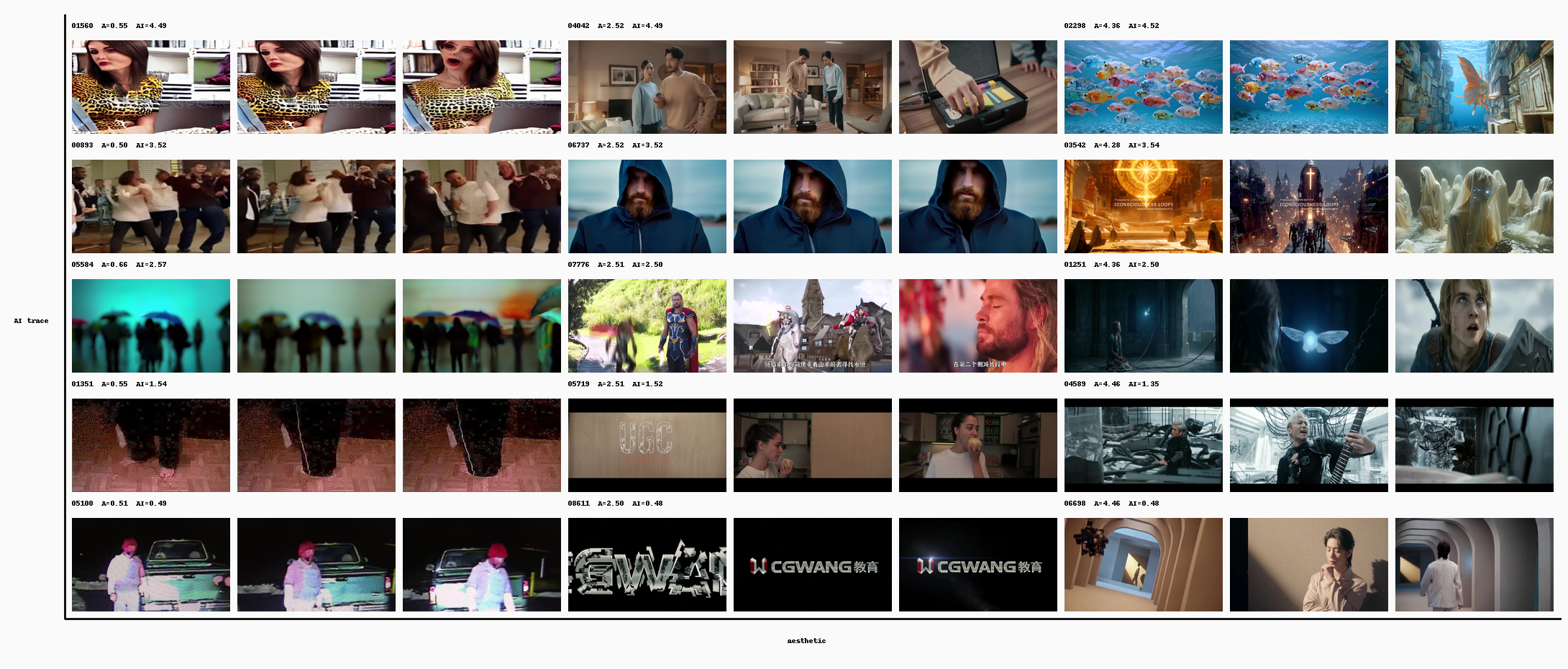}
    \caption{Visualization for examples of multiple AI trace-Aesthetic dimensions levels}
    \label{fig:aitrace-aesthetic}
\end{figure*}

Fig.~\ref{fig:aitrace-technical} and~\ref{fig:aitrace-aesthetic} provide additional visualizations of the relationships between AIGC-trace levels and technical and aesthetic quality scores for all the videos used in annotation.

\subsection{More Details for the Annotated Labels of TREND-10K}
\label{sec:Annotation}
\begin{figure*}[h!]
    \centering
    \includegraphics[width=0.65\linewidth]{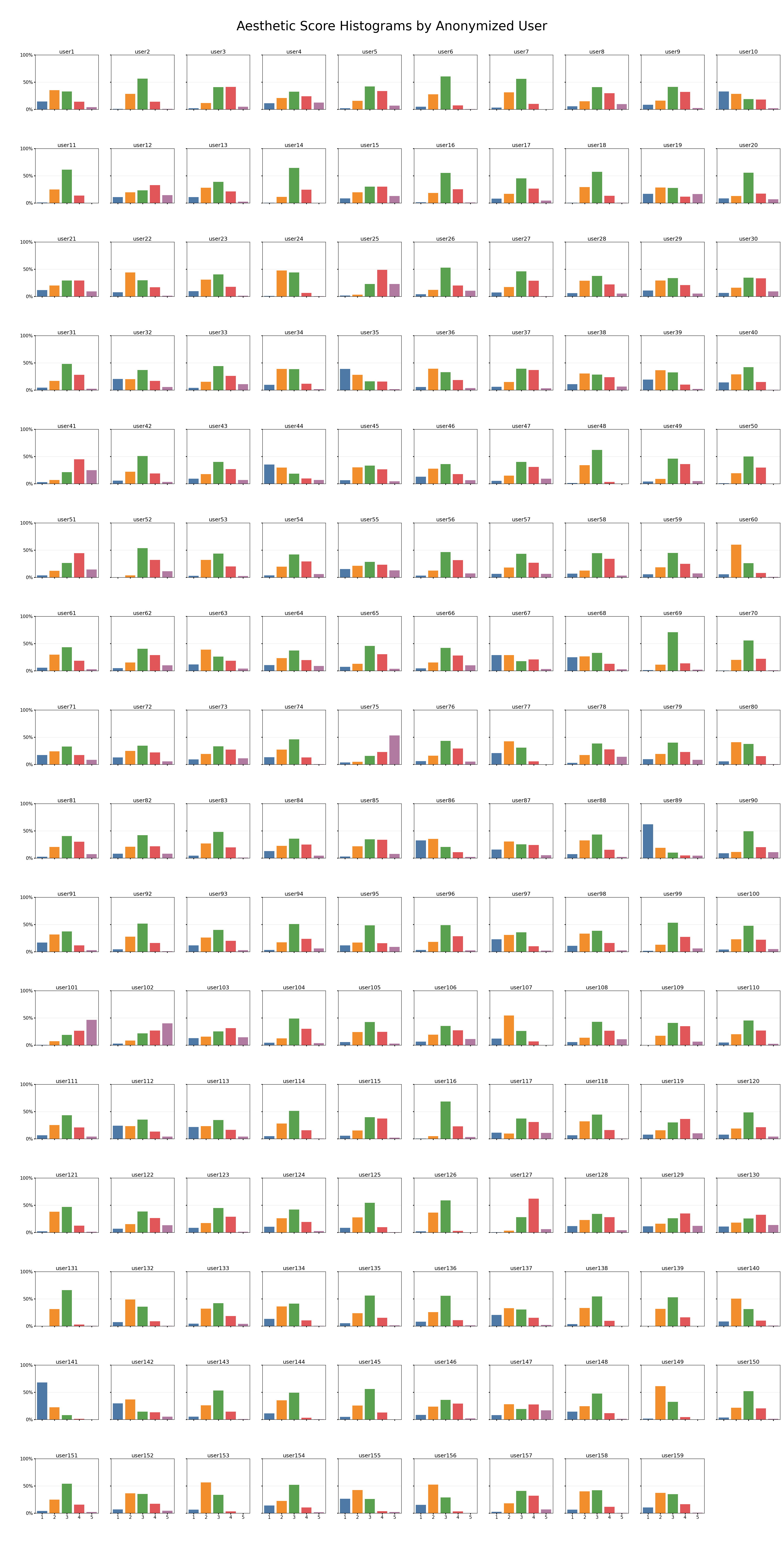}
    \caption{Aesthetic score distribution for all annotators}
    \label{fig:annotator_dist_aesthetic}
\end{figure*}

\begin{figure*}[h!]
    \centering
    \includegraphics[width=0.65\linewidth]{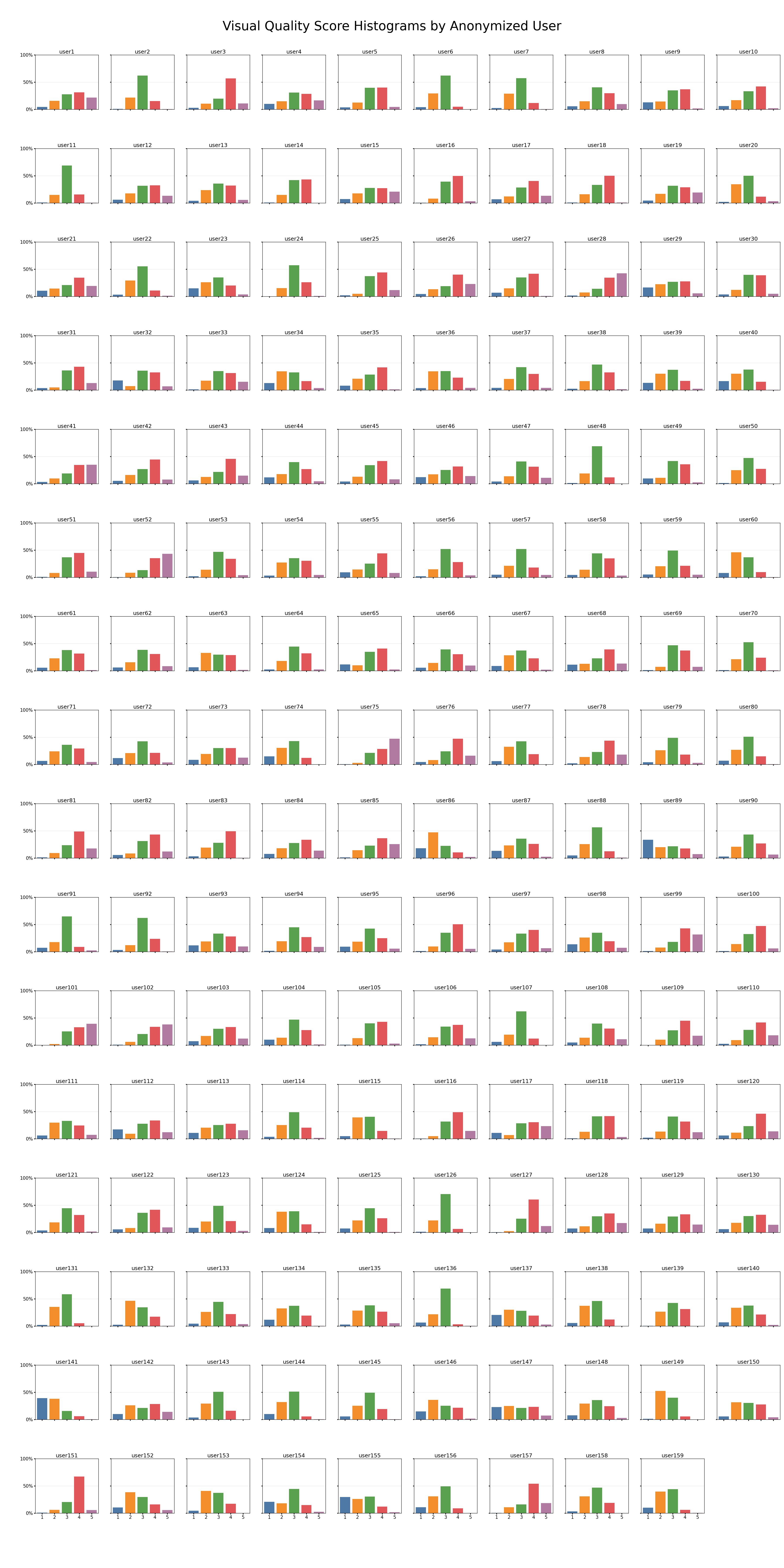}
    \caption{Technical score distribution for all annotators}
    \label{fig:annotator_dist_technical}
\end{figure*}

\begin{figure*}[h!]
    \centering
    \includegraphics[width=0.65\linewidth]{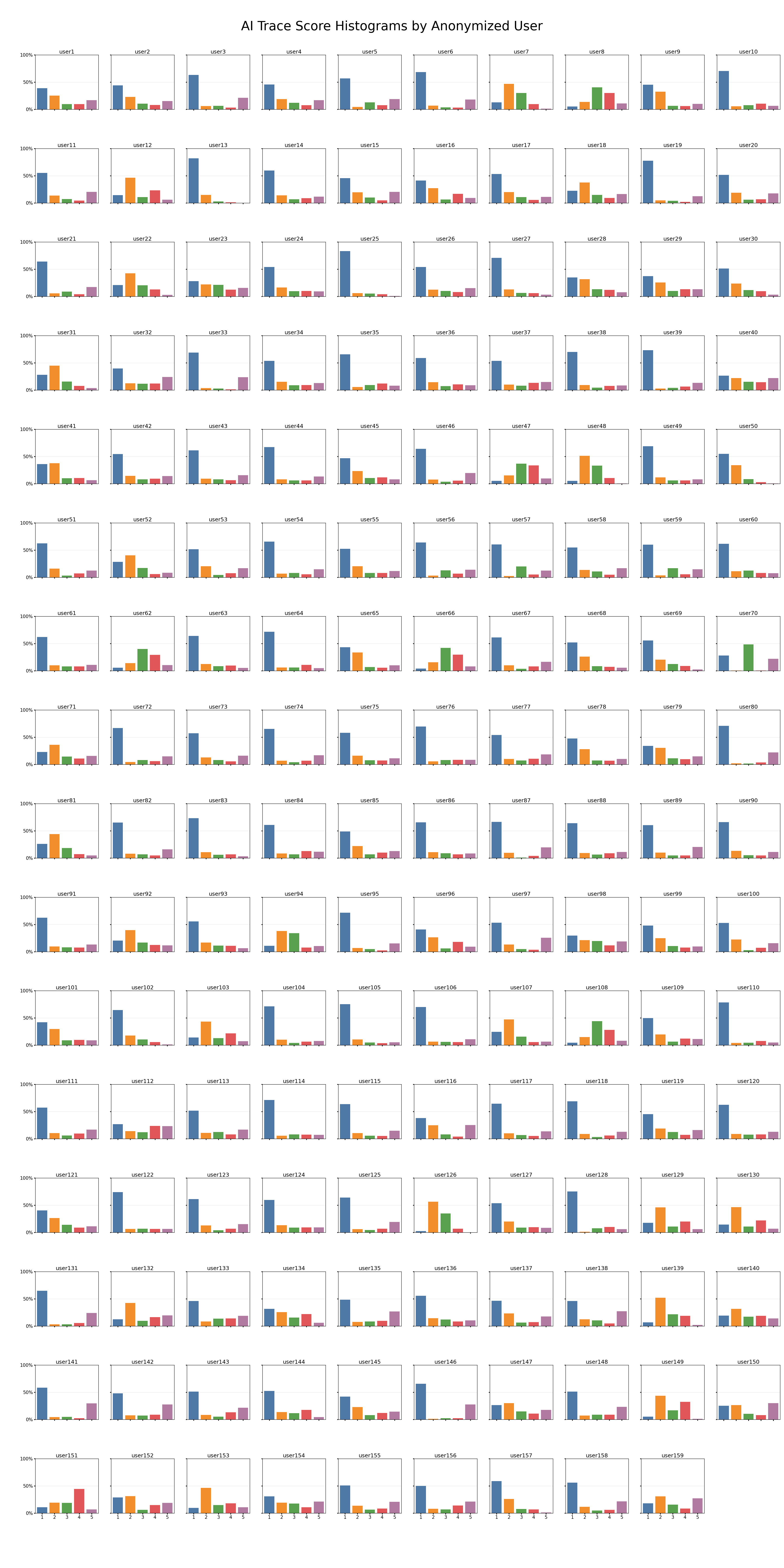}
    \caption{AIGC-trace score distribution for all annotators}
    \label{fig:annotator_dist_aitrace}
\end{figure*}

TREND-10K recruits $180$ subjects for subjective annotation. After strict quality screening, annotations from $21$ subjects are discarded. We then present the score distributions of the remaining $159$ annotators across the three evaluation dimensions within their assigned groups.
Fig.~\ref{fig:annotator_dist_aesthetic}, ~\ref{fig:annotator_dist_technical} and ~\ref{fig:annotator_dist_aitrace} present the score distributions of the retained annotators across the three evaluation dimensions, providing additional insights into the consistency and diversity of the collected annotations.
\newpage
\newpage
\newpage



\end{document}